\documentclass[10pt,twocolumn,letterpaper]{article}
\usepackage{iccv}            
\usepackage{graphicx}
\usepackage{amsmath}
\usepackage{amssymb}
\usepackage{booktabs}

\newcommand{\vx}{\mathbf{x}}
\newcommand{\vy}{\mathbf{y}}

\newcommand{\mbf}[1]{\mathbf{#1}}
\usepackage{amsmath,xfrac,array}

\usepackage[normalem]{ulem} 
\usepackage{booktabs}       
\usepackage{amsfonts}       
\usepackage{nicefrac}       
\usepackage{microtype}      
\usepackage{colortbl}         
\usepackage{amsmath}
\usepackage{bm}
\usepackage{graphicx}
\usepackage{graphbox}
\usepackage{svg}
\usepackage{subcaption}
\usepackage{multirow, makecell}
\usepackage{rotating}
\usepackage{caption}
\usepackage{wrapfig}
\usepackage{float}
\usepackage{xspace}
\usepackage{pifont}

\newcommand{\cmark}{\ding{51}}  
\newcommand{\xmark}{\ding{55}}  

\definecolor{Gray}{gray}{0.9}
\definecolor{lightbrown}{rgb}{0.827,0.784,0.667}

\usepackage{graphicx}
\usepackage{caption}
\usepackage{subcaption}
\usepackage{pgffor} 
\usepackage{lipsum} 
\usepackage{geometry} 

\usepackage{enumitem}
\newenvironment{tight_itemize}{
	\begin{itemize}[leftmargin=10pt]
		\setlength{\topsep}{0pt}
		\setlength{\itemsep}{0pt}
		\setlength{\parskip}{0pt}
		\setlength{\parsep}{0pt}
	}{\end{itemize}}

\definecolor{iccvblue}{rgb}{0.21,0.49,0.74}
\usepackage[pagebackref,breaklinks,colorlinks,allcolors=iccvblue]{hyperref}

\title{
Learning Visual Hierarchies in Hyperbolic Space for Image Retrieval
}

\author{
  Ziwei Wang$^1$$^2$$^*$ \quad
  Sameera Ramasinghe$^1$$^*$\quad
  Chenchen Xu$^1$\quad \\
  Julien Monteil$^1$\quad
  Loris Bazzani$^1$\quad
  Thalaiyasingam Ajanthan$^1$$^*$ \\ \\
  $^1$Amazon \quad
  $^2$Australian National University \quad
}

\begin{document}
\maketitle

\begin{abstract}
	Structuring latent representations in a hierarchical manner enables models to learn patterns at multiple levels of abstraction. However, most prevalent image understanding models focus on visual similarity, and learning visual hierarchies is relatively unexplored. In this work, for the first time, we introduce a learning paradigm that can encode user-defined multi-level complex visual hierarchies in hyperbolic space without requiring explicit hierarchical labels. As a concrete example, first, we define a part-based image hierarchy using object-level annotations within and across images. Then, we introduce an approach to enforce the hierarchy using contrastive loss with pairwise entailment metrics. Finally, we discuss new evaluation metrics to effectively measure hierarchical image retrieval. Encoding these complex relationships ensures that the learned representations capture semantic and structural information that transcends mere visual similarity. Experiments in part-based image retrieval show significant improvements in hierarchical retrieval tasks, demonstrating the capability of our model in capturing visual hierarchies. 
	\vspace{-3ex}

    {\let\thefootnote\relax\footnote{
    {$^*$ work done while at Amazon 
    }}}

\end{abstract}

\begin{figure}[t]
	\centering
	\includegraphics[width=0.95\linewidth]{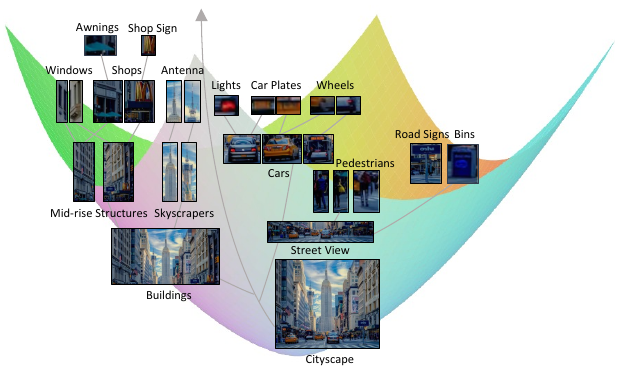}
	\caption{\textbf{An illustration of part-based image hierarchy organized in hyperbolic space.} At the highest level, we see the urban environment, composed of buildings, streets, and sky. 
		Zooming in, we find the building category, which further divides into skyscrapers, mid-rise structures, and more. 
		Each of them has its own visual elements, which in turn can be decomposed into sub-elements. 
        }
	\label{fig:space demo}
	\vspace{-4ex}
\end{figure}

\section{Introduction}
\label{sec:intro}
Humans organize knowledge of the world into hierarchies~\cite{nickel2017poincare} for efficient knowledge management. 
Developing models that encode such hierarchies is crucial for creating systems with holistic world understanding aligned with human perception. 
While this topic spans various modalities, we focus on encoding hierarchies in the visual domain. 
For many large image datasets, objects can be organized according to latent hierarchies~\cite{nickel2017poincare}, as evidenced by power law distributions~\cite{ravasz2003hierarchical}.
However, most prevalent image understanding models~\cite{schroff2015facenet, hoffer2015deep, he2016deep, chen2020simple, he2020momentum} focus on preserving visual similarity, and consequently, learning hierarchies in the visual domain remains relatively unexplored. 
These limitations hinder models' ability to generalize to tasks requiring hierarchical reasoning such as understanding scenes, object parts, and their interactions at multiple levels of abstraction.
We illustrate in Fig.~\ref{fig:space demo} an example that shows the complexity of a visual hierarchy where elements share similarities both within and across categories. 

Recent works have demonstrated the utility of hyperbolic representation spaces for capturing hierarchical relationships in an unsupervised setting~\cite{desai2023hyperbolic, ramasinghe2024accept,qiu2024hihpq}.
This \emph{emergent} latent structure is appealing; nevertheless, meaningful hierarchies are often task and data dependent, and aligning model behavior with such human-defined hierarchies is essential for many applications. 
To this end, we introduce a learning paradigm that can encode multi-level hierarchies as entailment pairs in hyperbolic space. 
As a concrete example, we first define a general part-based image hierarchy using object and part level classification annotations within and across images. Then, we introduce a model capable of structuring the latent space to preserve the defined hierarchy using only image/object level information. To our knowledge, we are the first to encode multi-level complex visual hierarchies without relying on explicit hierarchical labels or additional modalities. Finally, we introduce an evaluation metric to effectively measure hierarchical image retrieval.

Note that, hierarchy is an asymmetric relationship and has a high branching factor (see Fig.~\ref{fig:space demo}). To this end, we adopt the hyperbolic geometry as it provides a continuous approximation of such tree-like structures~\cite{nickel2017poincare, sala2018representation}. 
To enforce the hierarchy, we break it into pairwise entailment relationships between images, objects, and parts, at multiple levels within an image as well as across images at category level.
For pairwise entailment (asymmetric), we adapt the recently proposed angle-based asymmetric distance metric~\cite{ramasinghe2024accept} within the contrastive learning paradigm, and extend it to handle cases with multiple positive relationships. In contrast to symmetric distances such as the inner-product used in~\cite{nickel2018learning, khrulkov2020hyperbolic, ge2023hyperbolic}, this angle-based distance offers an additional degree of freedom along the radial axis to form emergent structures in the latent space.

For experimentation, we build a dataset of visual hierarchies using the bounding box annotations of OpenImages~\cite{kuznetsova2020open}. Our dataset includes entailment relationships between scenes, objects, and parts, within a single image as well as across images at the category level. Similarly, for hierarchical retrieval evaluation, we use the full training set to create ground truth hierarchy trees per scene/object. 
Additionally, we design a metric for evaluating hierarchical retrieval based on the optimal transport distance between the label distribution of the retrieval set and ground truth label distribution within the hierarchy tree. Combined with Recall@k metrics, this demonstrates that our method captures semantic and structural information, transcending mere visual similarity. 
Furthermore, to the best of our knowledge, our model is the first to generalize to out-of-domain image hierarchies, achieving strong performance on unseen and out-of-domain datasets.

Our contributions can be summarized as follows:
\begin{tight_itemize}
	\item To our knowledge, for the first time, we introduce a new learning paradigm to effectively encode user-defined multi-level complex visual hierarchies in hyperbolic space that does not require explicit hierarchical labels. 
	
	\item We adapt a contrastive loss using hyperbolic angle-based distance metric to enforce pairwise entailment relationships, and empirically demonstrate that pairwise entailment is sufficient to learn complex visual hierarchies.
	\item We introduce an optimal transport based evaluation metric to measure hierarchical image retrieval performance.
	\item We demonstrate superior generalization capabilities of our model beyond the user-defined hierarchies via out-of-domain unseen data evaluation and ablation experiments.
\end{tight_itemize}


\section{Preliminaries}\label{sec: lorentz}
We briefly review essential concepts in hyperbolic geometry here. We refer the reader to 
~\cite{ratcliffe1994foundations} for a comprehensive treatment.
Hyperbolic spaces are Riemannian manifolds with constant negative curvature and are fundamentally different from Euclidean or spherical space which has zero or constant positive curvature, respectively. The negative curvature enables properties such as divergence of parallel lines and exponential volume growth with radius~\cite{bridson2013metric}.
This volume growth property makes hyperbolic space an ideal candidate for embedding hierarchical and graph structured data~\cite{nickel2017poincare}, and has found many machine learning applications.

\vspace{-1ex}
\paragraph{Lorentz Model.}
The Lorentz model is a way to represent a hyperbolic space. It embeds the $d$-dimensional hyperbolic space \(\mathbb{H}^d\) with curvature $c$ within an $(d+1)$-dimensional Minkowski space \(\mathbb{R}^{d,1}\), a pseudo-Euclidean space with one negative dimension, as follows:
\begin{equation}
	\mathbb{H}^d = \left\{\mbf{x}\in\mathbb{R}^{d,1}\mid\langle \mbf{x}, \mbf{x} \rangle_{\mathbb{H}}=-1/c, x_0 > 0\right\}\ ,
\end{equation}
where the Lorentzian inner product is defined as,
\vspace{-1ex}
\begin{equation}
	\langle \mbf{x}, \mbf{y} \rangle_{\mathbb{H}} = -x_0y_0 + \sum_{i=1}^{d} x_iy_i\ .
	\vspace{-1ex}
\end{equation}
Here, the $0$-th dimension of the vector is treated as the time component and the rest as the space component.
From the definition of \(\mathbb{H}^d\), the time component can be written using the space component as follows:
\vspace{-1ex}
\begin{equation}
	\label{eq:time}
	x_{\text{time}} = x_0 = \sqrt{1/c  + \|\mbf{x}_{\text{space}}\|^2}\ ,
\end{equation}
\vspace{-1ex}
where $\|\cdot\|$ is the Euclidean norm and ${\mbf{x}_{\text{space}} = \mbf{x}_{1:d}}$.

\vspace{-1ex}
\paragraph{Tangent Spaces.}
The tangent space at a point \(\mbf{x}\in\mathbb{H}^d\) in hyperbolic space, is a Euclidean space that locally approximates hyperbolic space around \(\mbf{x}\). Exponential and logarithmic maps are used to project a point from a tangent space to hyperbolic space and vice versa.

\begin{figure*}[t]
	\centering
	\includegraphics[width=0.9\linewidth]{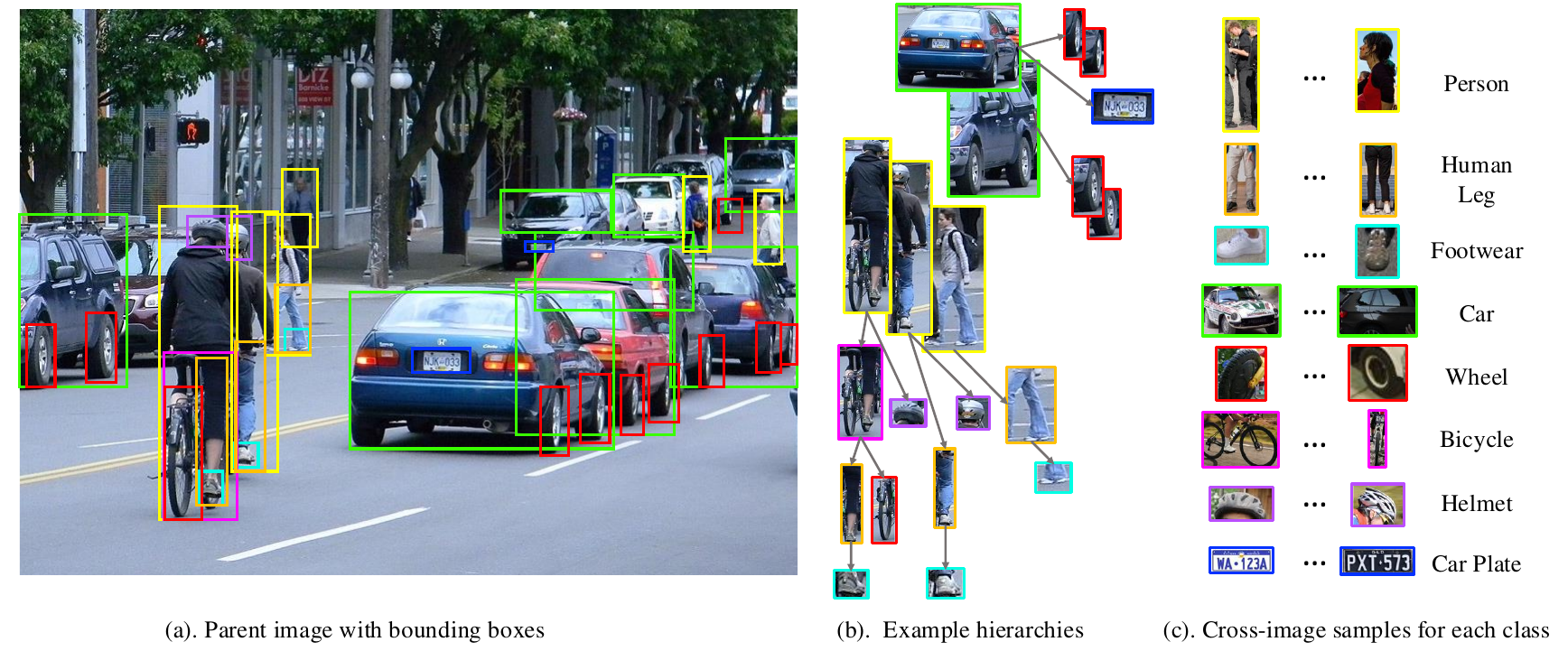}
	\caption{\textbf{An illustrative example image hierarchies.}
		a) Image $I$ with object-level bounding boxes. Each bounding box is entailed by $I$.
		b)  Hierarchies created via bounding box-to-bounding box entailment within $I$ (larger bounding boxes entail smaller ones).
		c) Cross-image hierarchy created by sampling $N$ bounding boxes with corresponding object classes from other images, which are then entailed by $I$. Find details of cross-image sampling in Sec.~\ref{sec: Part-Based Image Hierarchy}.
	}
	\label{fig:gen_data}
	\vspace{-3ex}
\end{figure*}

\section{Enforcing User-Defined Hierarchies}\label{sec:method}
Our aim is to define a hierarchy in images and enforce it in the latent space using hyperbolic geometry. We first define a part-based hierarchy in images, then discuss our approach to enforce it, and finally introduce our hierarchical retrieval metric.

\subsection{Part-Based Image Hierarchy}
\label{sec: Part-Based Image Hierarchy}

Visual hierarchies can be established in different ways, depending on the application. In this work, we are interested in a hierarchy that encapsulates the semantic relationship among objects in a scene. For this, {\em scene-object-part} hierarchy is appealing as it is useful for applications such as fine-grained object retrieval, object localization, and general scene understanding. This hierarchy has also been shown to emerge in hyperbolic image embeddings~\cite{khrulkov2020hyperbolic,ramasinghe2024accept}.

Given an image dataset with bounding box and object class annotations, we define a part-based image hierarchy where the full scene images -- typically containing multiple objects -- represent the highest level in the hierarchy, while individual objects constitute progressively lower levels, \emph{entailed} by the full image. In this, larger bounding boxes that significantly envelope smaller ones are considered to entail those smaller ones, establishing a nested hierarchy. 
In this way, from the full scene to the smallest bounding box, a hierarchy can be defined by recursively applying the entailment rule: {\em if $B$ contained in $A$, then $A$ entails $B$, denoted as $A\to B$}. 
An illustrative example is shown in Fig.~\ref{fig:gen_data}, where an example hierarchy could be \texttt{road scene $\to$ cyclist $\to$ bicycle $\to$ wheels}.

\vspace{-1ex}
\paragraph{Pairwise Entailment.}
\label{sec: Entailment pairing}
Our entailment rule above naturally facilitates a pairwise relationship.
Let $I\in\mathcal{I}$ be an image, either the full scene image or a cropped bounding box, 
and let $\mathcal{B}$ denote the set of all bounding boxes in the dataset.
Suppose $\mathcal{B}_I$ be the set of bounding boxes contained in the image $I$, \ie, $\mathcal{B}_I=\{b\in\mathcal{B} \mid b\subset I\}$\footnote{We use the $\subset$ notation to denote {\em contained in} relationship. In the case where $I$ is a cropped bounding box, this relationship is defined to hold if the majority (\eg., 80\%) of the small bounding box $b$ is contained in $I$.}. 
Note, each bounding box has an associated object label denoted by $b_l\in\mathcal{L}$.

We define the following pairwise entailment relationship: 
\vspace{-1ex}
\begin{equation}\label{eq:entail1}
	\operatorname{if}\ \  b \in \mathcal{B}_I,\ \  \operatorname{then}\ \  I \to b\ .   
\end{equation}
By applying this recursively, a tree-like hierarchy can be formed as shown in Fig.~\ref{fig:gen_data}b.
Note that our model can encode any ``user-defined hierarchy" represented as entailment pairs in Eq.~\eqref{eq:entail1}. Part-based image hierarchy is one use case.

Furthermore, if $I$ is a full scene image, we define an additional entailment relationship across images at the object level. Specifically, for each bounding box $b$ in the image $I$, we sample $K$ bounding boxes from other images with the same label $b_l$ and enforce entailment with the image $I$. Formally, let $a\in \mathcal{L}$, and let $\mathcal{B}^a_{I,K} \sim \{b\in B\mid b \not\subset I, b_l = a\}$ be the set of $K$ bounding boxes of label $a$ on images other than $I$. We enforce the entailment as follows:
\begin{equation}\label{eq:entail2}
	\operatorname{for} \operatorname{all}\ \  b \in \mathcal{B}_I,\ \  \operatorname{if}\ \  x \in \mathcal{B}^{b_l}_{I,K},\ \  \operatorname{then}\ \  I \to x\ .
\end{equation}
This additional entailment across images reinforces the semantic link between scenes and similar objects across different images (see Fig.~\ref{fig:gen_data}c).

We posit that these relationships help to structure a hierarchical understanding of images based on scene, object, and part relationships.

\vspace{-1ex}
\paragraph{Hierarchy Tree.}
As noted above, the part-based hierarchy forms a tree structure, where an image or cropped bounding box can be traversed using pairwise entailment relationships.
For evaluating hierarchical image retrieval, we construct this hierarchy tree per scene/object automatically using the full training set. 
However, the model is trained solely on pairwise entailment relationships and does not use the hierarchy tree.

\subsection{Angle-Based Entailment Loss}
We require an asymmetric distance function to enforce pairwise entailment relationships in hyperbolic space. To this end, we adapt the recently proposed hyperbolic-angle-based entailment loss~\cite{ramasinghe2024accept}, a smooth contrastive variant of the entailment cone loss~\cite{ganea2018entail}. 
The angle-only loss, 
without distance constraints on image pairs, provides flexibility to be distributed along the radial axis, allowing embeddings to align with the tree structure while preserving pairwise entailment.

\begin{figure}[t]
\centering
\includegraphics[width=0.95\linewidth]{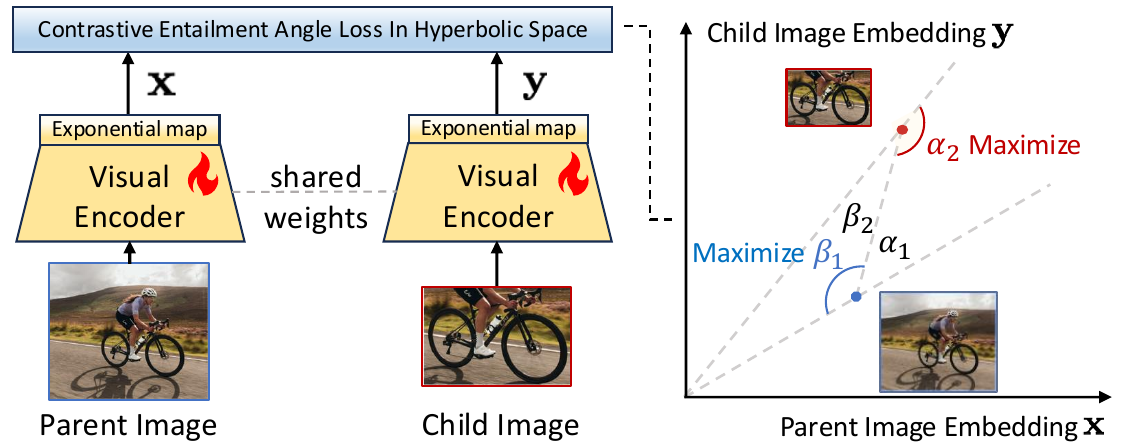}
\caption{\textbf{Learning multi-level hierarchies via contrastive entailment angle loss.}
	Our model first encodes parent-to-child pairs into embeddings with exponential mapping, then 
	maximizes $\beta_1$ and $\alpha_2$ using our contrastive entailment angle loss in hyperbolic space.
}
\label{fig:pipeline}
\vspace{-3ex}
\end{figure}

Our loss is a bidirectional version of~\cite{ramasinghe2024accept}, as illustrated graphically in Fig.~\ref{fig:pipeline}. In particular, given embeddings of an entailment pair, $\vx, \vy \in \mathbb{R}^d$ 
in the tangent space, where $\vx$ entails $\vy$, we maximize the angles $\beta_1$ and $\alpha_2$. This enforces entailment in a bidirectional manner. The angles $\beta_1$ and $\alpha_2$ can be computed using the exterior angle as follows:
\vspace{-1ex}
\begin{align}
\beta_1(\vx, \vy) &= \pi - \operatorname{ext}(\vx, \vy)\ ,\\\nonumber
\alpha_2(\vy, \vx) &= \operatorname{ext}(\vy, \vx)\ ,
\end{align}
where $\operatorname{ext}(\vx, \vy)$ is the exterior angle between $\vx$ and $\vy$ and takes the following form~\cite{desai2023hyperbolic}:
\vspace{-1ex}
\begin{equation}
\text{ext}(\mbf{x}, \mbf{y}) = \cos^{-1} \left( \frac{y_\text{time} + x_\text{time}\,c \langle \mbf{x}, \mbf{y} \rangle_{\mathbb{H}} }{\|\mbf{x}_{\text{space}}\|\sqrt{\left(c \langle \mbf{x}, \mbf{y}\rangle_{\mathbb{H}}\right)^2 -1}} \right)\ .
\end{equation}

In contrast to~\cite{ramasinghe2024accept}, in our case an embedding can belong to multiple entailment pairs in a batch. This corresponds to a case of multiple positives in the contrastive loss. Thus, we employ a multi-positive variant of the InfoNCE loss~\cite{oord2018representation} to align all entailment pairs while pushing apart the rest of the negative pairs. Precisely, let $\mathcal{D}=\{(\vx_i, \vy_i)\}$ be a batch of entailment pairs. 
For each parent embedding $\vx_i$, let $\mathcal{P}_i$ denote the set of child embeddings in the batch that have an entailment relationship with $\vx_i$, and let $\mathcal{A}$ denote the set of all candidate embeddings in the batch. Then the InfoNCE loss for parent-to-child can be written as:
{\footnotesize
\begin{align}
L^{p \to c} (\mathcal{D}, \kappa)
= -\mathbb{E}_{\mathcal{D}}\!\left[
\log \frac{
\sum\limits_{\vy \in \mathcal{P}_i}
\exp\!\left(\frac{\kappa(\vx_i, \vy)}{\tau}\right)
}{
\sum\limits_{\vy \in \mathcal{A}}
\exp\!\left(\frac{\kappa(\vx_i, \vy)}{\tau}\right)
}
\right]\ ,
\end{align}}
where $\kappa: \mathbb{R}^{d} \times \mathbb{R}^{d} \rightarrow \mathbb{R}$ is the similarity function, and $\tau$ is a learnable temperature parameter initialized to 0.07 following~\cite{desai2023hyperbolic}. Now, our bidirectional entailment loss can be written as:
\begin{equation}
\label{eq: angle loss}
L_{\text{angle}}(\mathcal{D}) = L^{p \to c} (\mathcal{D}, \beta_1) + L^{c \to p} (\mathcal{D}, \alpha_2)\ .
\end{equation}
Here, the similarity function $\kappa$ is replaced with angles $\beta_1$ and $\alpha_2$ so that the contrastive loss maximizes angles $\beta_1$ and $\alpha_2$ for matching entailment pairs in the batch $\mathcal{D}$.

In our implementation, we use a shared image encoder for both parent and child embeddings. Following the reparametrization of~\cite{desai2023hyperbolic}, we encode the space component of the Lorentz model in the tangent space at origin and project it onto the hyperboloid using the exponential map, enabling contrastive entailment angle loss computation in hyperbolic space.

\vspace{-1ex}
\paragraph{Loss in Euclidean Space.}
This entailment angle loss is general and can be effectively enforced in Euclidean space. In Euclidean space, the exterior angles are formulated as follows:
\vspace{-1ex}
\begin{equation}\label{eq:ext angle euc}
\text{ext}(\vx, \vy)_{\mathbb{E}} = \cos^{-1} \left( \frac{\lVert \vy \rVert^2 - \lVert \vx \rVert^2 - \lVert \vx-\vy \rVert^2}{2 \lVert \vx \rVert \cdot \lVert \vx-\vy \rVert} \right)\ .
\end{equation}
Now, the loss can be analogously derived.

\subsection{Hierarchical Retrieval Evaluation}\label{sec:ot-metric}
To evaluate retrieval performance on the hierarchy tree, we also introduce a metric that captures the label distribution in the dataset. This is important as different labels can have different numbers of instances and the standard metrics such as Recall@k is agnostic to it.

Consider the parent-to-child relationship, and let $\mathcal{H}_I$ denote the hierarchy tree originating from the query image $I$, containing $m$ labels. Then, the labels in $\mathcal{H}_I$ are the ground truth labels for the hierarchical retrieval for image $I$. Now, let $\mathbf{h}_I\in\mathbf{R}^m$ be the precomputed label distribution of $\mathcal{H}_I$. Then, our optimal transport (1-D Wasserstein) distance between the retrieved label distribution $\mathbf{r}_I$ and $\mathbf{h}_I$ is:
\vspace{-1ex}
\begin{equation}
\operatorname{OT}(\mathbf{h}_I, \mathbf{r}_I) = \operatorname{Wasserstein}(\bar{\mathbf{h}}_I, \bar{\mathbf{r}}_I)\ ,
\end{equation}
where $\bar{\cdot}\in\mathbf{R}^{m+1}$ is the label distribution with {\em other} class added\footnote{For $\mathbf{h}_I$ other class has zero mass, and for $\mathbf{r}_I$ all labels not in $\mathcal{H}_I$ are combined to form the other class.}. Note that a smaller distance indicates better alignment.

\vspace{-1ex}
\section{Related Work}
Hyperbolic geometry allows for exponential volume growth with respect to the radius~\cite{bridson2013metric}, making it effective for embedding hierarchical structures. 
This advantage has led to significant research into leveraging hyperbolic representations for various data types,
including molecular structures~\cite{yu2022skin}, 3D~\cite{chen2020learning, hsu2021capturing, wang2023hypliloc}, images~\cite{khrulkov2020hyperbolic, guo2022clipped, ge2023hyperbolic, kim2023hier,qiu2024hihpq}, text data~\cite{tifrea2018poincar, tifrea2018poincar, ganea2018hyperbolic, zhu2020hypertext,jeong2024simple}, and vision-language data~\cite{desai2023hyperbolic, ramasinghe2024accept, kong2024hyperbolic, alper2024hierarcaps}.

Hyperbolic embeddings 
can be learned through standard deep learning layers~\cite{khan2022transformers} with hyperbolic projection~\cite{nickel2017poincare} or using hyperbolic neural networks~\cite{ganea2018hyperbolic}.
Many prior NLP, computer vision and knowledge graph studies learn hierarchies from partially order data~\cite{ganea2018hyperbolic, vilnis2018probabilistic, liu2019hyperbolic}, or minimizing geodesic distance or maximizing similarities~\cite{nickel2017poincare,nickel2018learning,yan2021unsupervised,ge2023hyperbolic}.
Ganea et al.~\cite{ganea2018hyperbolic} introduced an angle-based entailment cone loss which pushes child nodes into the cone emanating from the parent node embedding.
This approach has been applied to both text~\cite{ganea2018hyperbolic} and image data~\cite{dhall2020hierarchical} with label hierarchies.
Recently, this hyperbolic entailment loss was adapted
for contrastive learning to develop representations in vision-language models~\cite{desai2023hyperbolic, ramasinghe2024accept,alper2024hierarcaps, pal2024compositional}.
However, such methods remain relatively unexplored in the image domain.
We adapted the angle-based entailment loss from~\cite{ramasinghe2024accept} to encode part-based image hierarchies.
Many previous works are limited to learning hierarchies for single-class images using predefined labeled hierarchies, such as ImageNet~\cite{liu2020hyperbolic, khrulkov2020hyperbolic} or hand-labeled data~\cite{dhall2020hierarchical}.
In this work, we propose a learning method that fine-tunes pre-trained models on large-scale datasets for \textbf{general images} (with multiple classes per image) \textbf{without hierarchical labels}.
The most relevant approach, HCL~\cite{ge2023hyperbolic}, 
models simple scene-to-object hierarchies, whereas we capture more complex, multi-level part-based hierarchies directly from image data, extending beyond visual similarity.

\vspace{-1ex}
\section{Experiments}
\label{sec:experiment}
\vspace{-1ex}
\paragraph{Datasets.}
For training and evaluation, we construct \textit{HierOpenImages}, a novel dataset containing pairwise part-based image hierarchies built from the OpenImages dataset~\cite{kuznetsova2020open}.
We further evaluate generalization on out-of-domain datasets and hierarchies.

\vspace{-1ex}
 \paragraph{Models.}We evaluate the performance of learning multi-level image hierarchies using two popular visual encoders 1) CLIP ViT (B/16)~\cite{radford2021learning}, pretrained on large-scale image-text pairs from the Internet,
and 2) MoCo-v2 (ResNet-50)~\cite{chen2020improved}, pretrained on ImageNet.
Note that both CLIP and MoCo-v2 models are fine-tuned and evaluated in an \textbf{image encoder only} setting.
We use these pretrained image models as our baseline and compare our proposed angle-based hyperbolic method against its Euclidean counterpart.
Each model is fine-tuned on \textit{HierOpenImages} using the proposed contrastive angle-based entailment loss, and $\dagger$ denotes fine-tuning.
Note that $*$ denotes the gated entailment angle metric, which first uses the entailment angle as a cone-shaped filter and then applies the pretrained metric within the gated set for fine-grained ranking (see Supp. \S \ref{sec:metrics}).

We compare the state-of-the-art hyperbolic hierarchical image-only model HCL~\cite{ge2023hyperbolic}, which is trained on scene-to-object relationships from the OpenImages dataset~\cite{kuznetsova2020open} in both hyperbolic and Euclidean space.
Accordingly, we evaluate the retrieval performance of HCL~\cite{ge2023hyperbolic} using both hyperbolic distance and cosine similarity.
For completeness, we also fine-tuned HCL$^\dagger$~\cite{ge2023hyperbolic} on the \textit{HierOpenImages} dataset.
Note that fine-tuning or evaluating with a text encoder is not possible on the \textit{HierOpenImages} dataset.
Furthermore, previous image-only supervised methods, trained on predefined single-class image hierarchies~\cite{dhall2020hierarchical, khrulkov2020hyperbolic} are not directly comparable. 
These methods require all image classes to be predefined during training,  which is not feasible for complex multi-object scenes in the OpenImages~\cite{kuznetsova2020open} and LVIS~\cite{gupta2019lvis} datasets, where images contain multiple objects with diverse class labels.
\footnote{Updated results after fixing an extra normalization issue in the code which impacted the results of all methods; overall trends and conclusions remain unaffected.
}

\subsection{Main Results}

\label{sec: Part-based Image Retrieval}
In same-class retrieval, we assess whether the retrieved image belongs to the \emph{same class} as the query image. In hierarchical retrieval, we verify if the retrieved image exists within the \emph{hierarchy tree} of the query image, specifically evaluating the quality of the learned hierarchical representations.

\vspace{-1ex}
\paragraph{Same-Class Retrieval.}
Denoting the full image as \emph{parent} and the bounding box as \emph{child}, we evaluate retrieval tasks in both child-to-parent and parent-to-child directions.
Table~\ref{tab:retrieval_1} shows the retrieval accuracy of $\text{top-k}=\{5, 10, 50, 100\}$.
We notice a consistent improvement with our proposed hyperbolic model, across all metrics and model variants.
This highlights the relevance of our angle-based entailment loss and the advantages of learning hierarchical image embeddings in hyperbolic space.

\vspace{-1ex}

\begin{table}[t]
    \centering
    \small
    \vspace{-1ex}
    \resizebox{0.95\columnwidth}{!}{%
    \begin{tabular}{cc c cc cc}

        \toprule
         Vision  Encoder & Model & Metrics & \multicolumn{1}{c}{\cellcolor{orange!10} Top-5} & \multicolumn{1}{c}{\cellcolor{orange!20}Top-10} &  \multicolumn{1}{c}{\cellcolor{orange!30} Top-50} & \multicolumn{1}{c}{\cellcolor{orange!40}Top-100 }  \\ \midrule

        \rowcolor{Gray}\multicolumn{0}{c}{\textbf{Child-to-Parent }} & & & & & & \\
    
         \multirowcell{3}{ CLIP ViT} & \multirow{1}{*}{CLIP} & Cos Sim. & 53.04 & 51.69 & 47.87 & 45.79 \\
         & \multirow{1}{*}{CLIP-euc$^\dagger$} & Euc Ang$.*$  &  75.63 & 74.65 & 72.25 &  70.52 \\
         & \multirow{1}{*}{\cellcolor{blue!8}
         CLIP-hyp$^\dagger$} & \cellcolor{blue!8} Hyp Ang$.*$ & \cellcolor{blue!8} \textbf{77.28}  & \cellcolor{blue!8} \textbf{75.91} & \cellcolor{blue!8} \textbf{72.85} & \cellcolor{blue!8} \textbf{70.94}  \\
         \midrule
         & \multirow{1}{*}{HCL}  & Hyp Dist.  & 46.57 & 44.86 & 39.64 & 36.61 \\
         & \multirow{1}{*}{HCL} & Cos Sim. & 55.48 & 54.81 & 52.23 & 50.67 \\
         & \multirow{1}{*}{HCL$^\dagger$} & Hyp Dist.  & 39.71 & 37.55 & 32.13 & 29.44 \\
         & \multirow{1}{*}{HCL$^\dagger$} & Cos Sim.  & 56.76 & 55.85 & 53.61 & 52.16 \\
         \multirowcell{1}{MoCo-v2} &  \multirow{1}{*}{MoCo} & Cos Sim. & 53.72 & 52.71 & 50.41 & 49.19 \\
         & \multirow{1}{*}{MoCo-euc$^\dagger$} & Euc Ang$.*$ & 66.78 & 66.58 & 65.42 & 64.40 \\
         & \multirow{1}{*}{\cellcolor{blue!8} MoCo-hyp$^\dagger$} & \cellcolor{blue!8} Hyp Ang$.*$  & \cellcolor{blue!8} \textbf{68.62} & \cellcolor{blue!8} \textbf{67.42} & \cellcolor{blue!8} \textbf{66.21} & \cellcolor{blue!8} \textbf{64.94} \\
        \midrule
          \rowcolor{Gray}{\textbf{Parent-to-Child }}  & & & & & &  \\
         
         \multirowcell{3}{CLIP ViT } & \multirow{1}{*}{CLIP} & Cos Sim. & 69.94  & 67.74 & 62.50  & 60.21 \\
         & \multirow{1}{*}{CLIP-euc$^\dagger$} & Euc Ang$.*$ & 73.74 & 72.57 & 70.04 & 68.78 \\
         & \multirow{1}{*}{\cellcolor{blue!8} CLIP-hyp$^\dagger$} & \cellcolor{blue!8} Hyp Ang$.*$  & \cellcolor{blue!8} \textbf{74.48} & \cellcolor{blue!8} \textbf{73.27} & \cellcolor{blue!8} \textbf{70.78} & \cellcolor{blue!8} \textbf{69.46} \\
         
         \midrule
         & \multirow{1}{*}{HCL}  & Hyp Dist.  & 50.86 & 49.81 & 47.34 & 46.06 \\
         & \multirow{1}{*}{HCL} & Cos Sim. & 53.64 & 53.04 & 51.34 & 50.59 \\
         & \multirow{1}{*}{HCL$^\dagger$} & Hyp Dist.  & 53.29 & 52.39 & 50.04 & 48.74 \\
         & \multirow{1}{*}{HCL$^\dagger$} & Cos Sim.  & 54.99 & 54.36 & 52.86 & 52.03 \\
         \multirowcell{1}{MoCo-v2} & \multirow{1}{*}{MoCo} & Cos Sim.  & 63.65 & 62.21 & 58.85 & 57.28 \\
         & \multirow{1}{*}{MoCo-euc$^\dagger$} & Euc Ang$.*$ & 68.87 & 67.94 & 65.64 & 64.42 \\
         & \multirow{1}{*}{\cellcolor{blue!8} MoCo-hyp$^\dagger$} & \cellcolor{blue!8} Hyp Ang$.*$ & \cellcolor{blue!8} \textbf{69.50} & \cellcolor{blue!8} \textbf{68.14} & \cellcolor{blue!8} \textbf{65.67} & \cellcolor{blue!8} \textbf{64.46} \\
         \bottomrule
         
    \end{tabular}
    }
    \caption{\textbf{Part-based same-class image retrieval evaluation.}
    For child-to-parent image retrieval, the retrieved parent must contain the object class of the query child. For parent-to-child image retrieval, the retrieved child must match a class within the parent. 
    $\dagger$ denotes models fine-tuned on the \textit{HierOpenImages} dataset.
    Our proposed method is shaded in purple.
    }
    \label{tab:retrieval_1}
    \vspace{-2ex}
\end{table}

\begin{table}[t]
    \centering
    \small
    \vspace{-1ex}
    \resizebox{0.9\columnwidth}{!}{%
    \begin{tabular}{cc cc c c}

        \toprule
          Metrics & Model & \multicolumn{1}{c}{Dist. Func.}  &  \multicolumn{1}{c}{Top-150k} &  \multicolumn{1}{c}{Top-200k} &  \multicolumn{1}{c}{Top-250k} \\ \midrule

          \rowcolor{Gray}{\textbf{CLIP ViT}}  & & & & &  \\

         \multirowcell{3}{Recall $\uparrow$} & \multirow{1}{*}{CLIP} & Cos Sim.  & 66.63 & 77.15  & 86.57 \\
         & \multirow{1}{*}{CLIP-euc$^\dagger$} & Euc Ang$.*$ & 76.46 & 85.08 & 91.38 \\
         & \multirow{1}{*}{\cellcolor{blue!8} CLIP-hyp$^\dagger$} & \cellcolor{blue!8} Hyp Ang$.*$ & \cellcolor{blue!8} \textbf{77.00} & \cellcolor{blue!8} \textbf{85.75} & \cellcolor{blue!8} \textbf{91.89}  \\
         
         \midrule

         \multirowcell{3}{OT Distance $\downarrow$} & \multirow{1}{*}{CLIP} & Cos Sim. & 21.31 & 22.74 & 23.79 \\
         & \multirow{1}{*}{CLIP-euc$^\dagger$} & Euc Ang$.*$ &  15.65 & 18.15 & 21.09\\
         & \multirow{1}{*}{\cellcolor{blue!8} CLIP-hyp$^\dagger$} & \cellcolor{blue!8} Hyp Ang$.*$  & \cellcolor{blue!8} \textbf{14.96} & \cellcolor{blue!8} \textbf{17.57} & \cellcolor{blue!8} \textbf{20.76} \\

         \midrule
         \rowcolor{Gray}{\textbf{MoCo-v2}}  & & & & &  \\

         \multirowcell{6}{Recall $\uparrow$} 
         & \multirow{1}{*}{HCL} & Hyp Dist. & 51.83 & 64.84 & 77.66 \\
         & \multirow{1}{*}{HCL} & Cos Sim.  & 57.05 & 67.65 & 78.71 \\
         & \multirow{1}{*}{HCL$^\dagger$} & Hyp Dist. & 54.51 & 66.70 & 78.83 \\
         & \multirow{1}{*}{HCL$^\dagger$} & Cos Sim. & 60.13 & 70.26 & 80.51 \\
         & \multirow{1}{*}{MoCo} & Cos Sim.  & 65.01 & 75.85 & 85.61 \\
         & \multirow{1}{*}{MoCo-euc$^\dagger$} & Euc Ang$.*$ & 74.84 & 83.71 & 90.46 \\
         & \multirow{1}{*}{\cellcolor{blue!8} MoCo-hyp$^\dagger$} & \cellcolor{blue!8} Hyp Ang$.*$  & \cellcolor{blue!8} \textbf{74.85} & \cellcolor{blue!8} \textbf{83.79} & \cellcolor{blue!8} \textbf{90.61}  \\
         
         \midrule

         \multirowcell{6}{OT Distance $\downarrow$} 
         & \multirow{1}{*}{HCL} & Hyp Dist. & 24.58 & 24.84 & 25.01 \\
         & \multirow{1}{*}{HCL} & Cos Sim.  & 23.58 & 24.65 & 25.11 \\
         & \multirow{1}{*}{HCL$^\dagger$} & Hyp Dist. & 25.30 & 25.63 & 25.53 \\
         & \multirow{1}{*}{HCL$^\dagger$} & Cos Sim. & 21.94 & 23.58 & 24.58 \\
         & \multirow{1}{*}{MoCo} & Cos Sim. & 18.60 & 20.79 & 22.56 \\
         & \multirow{1}{*}{MoCo-euc$^\dagger$} & Euc Ang$.*$ & 16.65 & 18.75 & 21.38  \\
         & \multirow{1}{*}{\cellcolor{blue!8} MoCo-hyp$^\dagger$} & \cellcolor{blue!8} Hyp Ang$.*$  & \cellcolor{blue!8} \textbf{16.59} & \cellcolor{blue!8} \textbf{18.65} & \cellcolor{blue!8} \textbf{21.29} \\
         \bottomrule
    \end{tabular}
    }
    \caption{\textbf{Part-based hierarchical evaluation of parent-to-child image retrieval on \textit{HierOpenImages}.
    }
    Results are evaluated using the ground truth hierarchy tree and the hierarchical distribution of the test set.
    Smaller OT distance indicates better distribution alignment.
    }
    \label{tab:retrieval_2}
    \vspace{-2ex}
\end{table}

\begin{figure}[t]
\vspace{-1ex}
\centering
\includegraphics[width=0.8\linewidth]{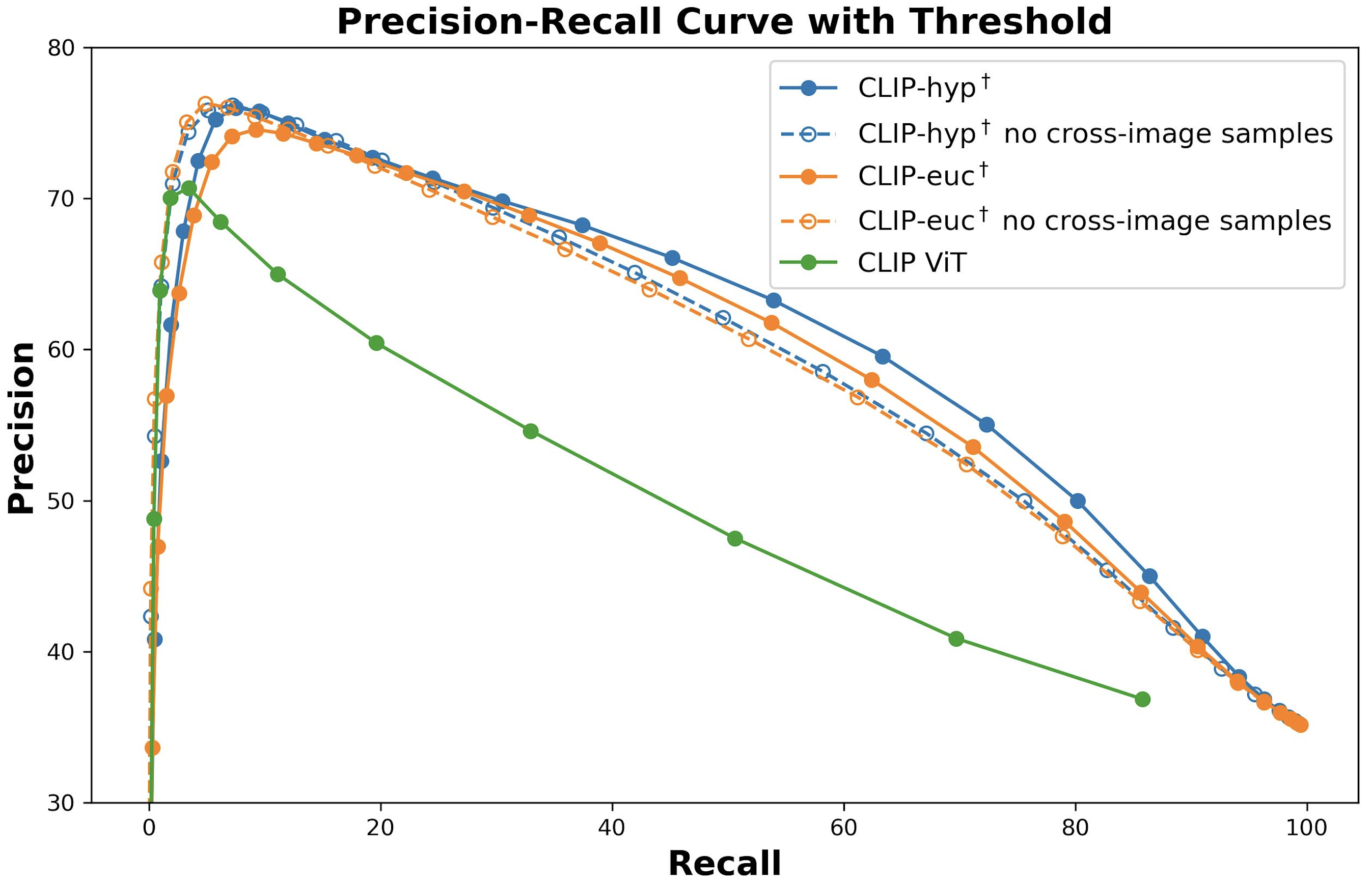}
    \vspace{-1ex}
\caption{
	\textbf{Precision-Recall curves of CLIP ViT models on hierarchical retrieval. }
	Dotted lines show models trained only on hierarchical entailment data within the same images; solid lines represent models trained with additional cross-image scene-to-object samples. 
}
\vspace{-2ex}
\label{fig:PR curve}
\end{figure}
\vspace{-1ex}
\paragraph{
Hierarchical Retrieval via the Learned Latent Space Distribution.}
Table~\ref{tab:retrieval_2} evaluates the hierarchical structure of the latent space by retrieving a large number of child images from parent images.
We use recall to measure the percentage of ground truth images that are successfully retrieved.
Moreover, we check the alignment between the retrieved distribution and the underlying hierarchical distribution of the full test set. 
Good distribution alignment is a desirable property for fine-grained retrieval as the \textit{retrieved set should capture the hierarchies present in the data distribution}. 
We propose to measure distribution alignment using the optimal transport (Wasserstein distance), with a smaller distance indicating a closer match (see Sec.~\ref{sec:ot-metric}).

As shown in Table~\ref{tab:retrieval_2}, our hyperbolic model better captures the hierarchical distribution of the test set compared to the Euclidean model, achieving better recall and OT distance in all cases.
All fine-tuned models using our angle-based entailment loss show clear improvement over the baseline models.
In contrast, HCL~\cite{ge2023hyperbolic} shows only a small gain after fine-tuning, suggesting that its training method is not well suited for capturing the complex visual hierarchies in our data.

\vspace{-1ex}
\paragraph{Effect of Enforcing Cross Image Entailment.} 
To compare the emergent behaviors of
the hyperbolic and Euclidean models, we trained the CLIP ViT model solely on hierarchical part-based entailment data within individual images (entailment pairs with high visual similarity), omitting any cross-image image-to-bounding-box samples.

Fig.~\ref{fig:PR curve} shows the Precision-Recall (PR) curve with increasing $\beta_1$ angle thresholds ($0$ to $\pi$) for CLIP-hyp$^\dagger$ and CLIP-euc$^\dagger$, and with cosine similarity thresholds (0 to 1) for the baseline CLIP.
Fine-tuning in both Euclidean and Hyperbolic spaces substantially improves performance from the pretrained model.
Introducing cross-image pairs further improves recall in the mid-to-high recall regions, indicating better alignment across semantically related but visually diverse examples, with only a small precision drop at the top ranks due to the added variability from cross-image relationships.

\begin{figure}[t]
\centering\includegraphics[width=1\linewidth]{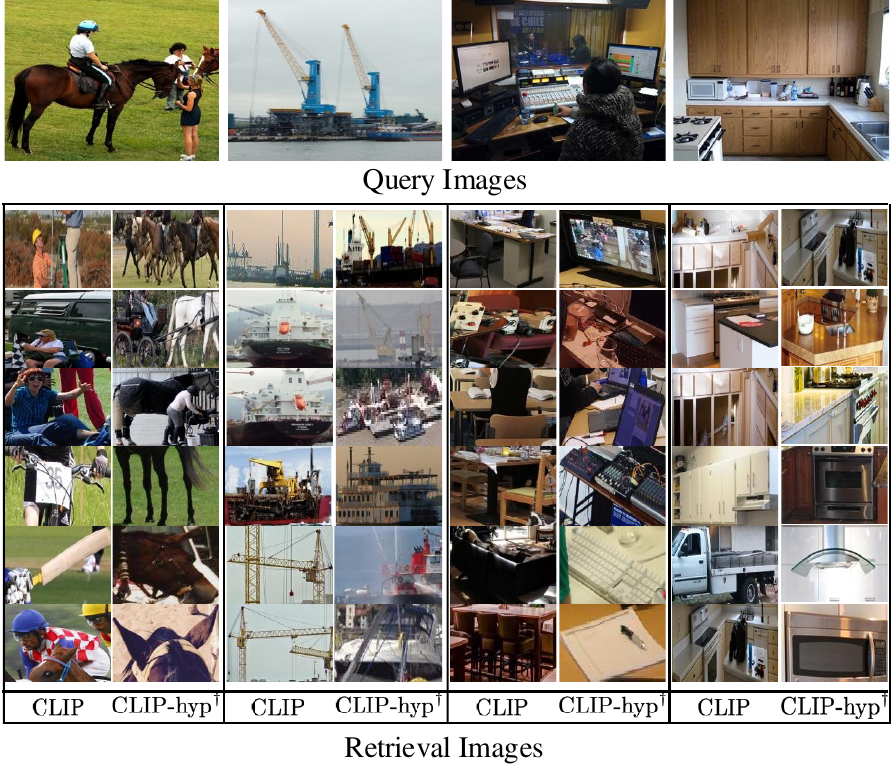}
\vspace{-3ex}
\caption{\textbf{Example of parent-to-child retrieval using CLIP ViT and our CLIP-hyp$^\dagger$ model.} 
	Results are ordered by ascending norms.
    Our model retrieves images matching the predefined {\em scene-object-part} hierarchy, placing high-level objects near the origin (\eg, harbor $\rightarrow$ boart parts), and grouping semantically related but visually distinct objects (\eg, microwave oven \& kitchen hood).  
}\vspace{-1ex}
\label{fig:demo retrieval}
\end{figure}
\vspace{-5pt}

\vspace{-1ex}

\paragraph{Qualitative Results. }
Following~\cite{desai2023hyperbolic, Ge2021RobustCL, ramasinghe2024accept}, we use parent-to-child image traversals with results ordered by increasing embedding norm, to illustrate the latent space, which:
(1) aligns with the predefined {\em scene-object-part} hierarchy, placing high-level objects near the origin.
(2). groups diverse objects under the same lower hierarchical branch, even if they are visually distinct (\textit{e.g.,} keyboard and monitor under the studio image).
Moreover, Figure~\ref{fig:dist} shows an example of the model's emergent structure where high-frequency, ambiguous image crops are naturally pushed toward the boundary of hyperbolic space.
Find details in supplementary materials.

\begin{figure}[H]
    \centering 
    \vspace{-2ex}
    \includegraphics[width=1\linewidth]{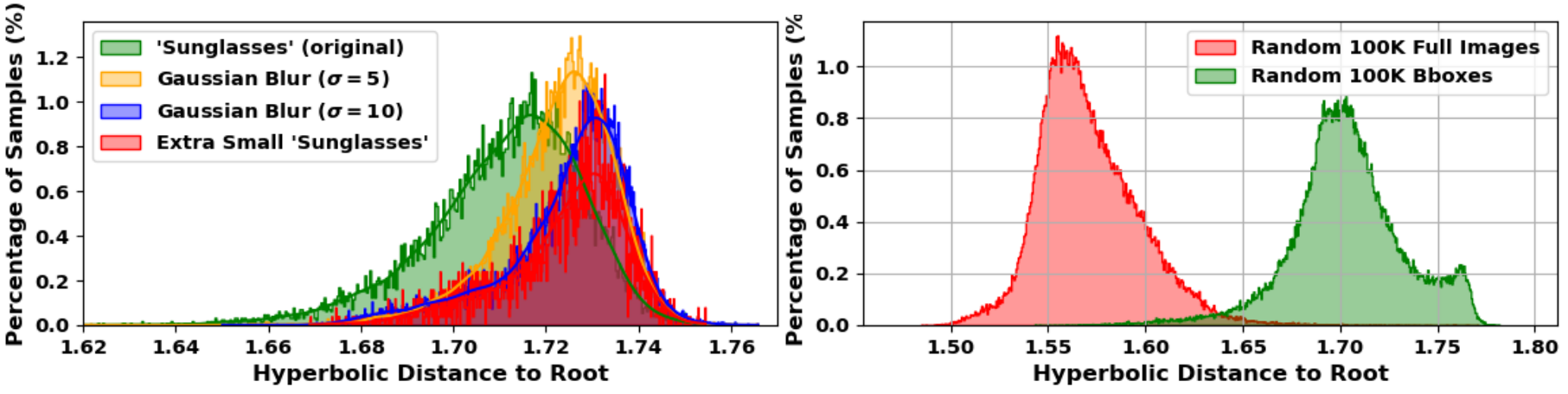} 
    \caption{\textbf{Example of the model's emergent structure.}
    }
    \label{fig:dist}
    \vspace{-2ex}
\end{figure}

\vspace{-1ex}
\subsection{
Generalization Evaluation
} 
We evaluate the generalization of our method through image retrieval on unseen datasets and hierarchies.
Similar to Table~\ref{tab:retrieval_1}-\ref{tab:retrieval_2}, Table~\ref{tab:unseen data} presents part-based same-class and hierarchical retrieval performance on the \textbf{out-of-domain} LVIS dataset~\cite{gupta2019lvis}, 
which contains 1,203 fine-grained object classes, including many rare objects absent from the training \textit{HierOpenImages} dataset.
Table~\ref{tab:coco_voc} evaluates image retrieval performance on VOC and COCO datasets following the zero-shot object detection in~\cite{pal2024compositional} where ground-truth boxes are used as region proposals, and then predict via top-5 majority vote.
Results demonstrate that our visual hierarchical learning significantly enhances model generalization on unseen datasets and image hierarchies.

\begin{table}[t]
    \centering
    \small
    \vspace{-1ex}
    \resizebox{0.85\columnwidth}{!}{%
    \begin{tabular}{c c cc cc}

        \toprule
         Model & Dist. Func. & \multicolumn{1}{c}{\cellcolor{orange!10} Top-5} & \multicolumn{1}{c}{\cellcolor{orange!20}Top-10} &  \multicolumn{1}{c}{\cellcolor{orange!30} Top-50} & \multicolumn{1}{c}{\cellcolor{orange!40}Top-100 }  \\ \midrule

        \rowcolor{Gray}\multicolumn{0}{c}{\textbf{Child-to-Parent }} & & & & & \\

         \multirow{1}{*}{CLIP} & Cos Sim. & 15.00 & 14.22 & 11.91 & 10.48 \\
         \multirow{1}{*}{CLIP-euc$^\dagger$} & Euc Ang$.*$  & 28.45 & 26.76 & 22.65 & 20.43  \\
         \multirow{1}{*}{\cellcolor{blue!8}
         CLIP-hyp$^\dagger$} & \cellcolor{blue!8} Hyp Ang$.*$ & \cellcolor{blue!8} \textbf{28.84}  & \cellcolor{blue!8} \textbf{27.02} & \cellcolor{blue!8} \textbf{22.87} & \cellcolor{blue!8} \textbf{20.53} \\
         \midrule

        \rowcolor{Gray}{\textbf{Parent-to-Child }}  & & & & &  \\
         
         CLIP & Cos Sim. & 28.37  & 27.35 & 24.99  & 23.36 \\
         \multirow{1}{*}{CLIP-euc$^\dagger$} & Euc Ang$.*$ & 28.84  & 27.02  & 22.65  & 20.43 \\
         \multirow{1}{*}{\cellcolor{blue!8} CLIP-hyp$^\dagger$} & \cellcolor{blue!8} Hyp Ang$.*$  & \cellcolor{blue!8} \textbf{30.27} & \cellcolor{blue!8} \textbf{29.53} & \cellcolor{blue!8} \textbf{27.83} & \cellcolor{blue!8} \textbf{26.81} \\
         
         \bottomrule
         \bottomrule
         \addlinespace[15pt]

         \toprule
         Metrics & Model & \multicolumn{1}{c}{Dist. Func.} & \multicolumn{1}{c}{ Top-10k} &  \multicolumn{1}{c}{Top-20k} &  \multicolumn{1}{c}{ Top-30k}  \\
         \midrule
          \multirowcell{3}{Recall $\uparrow$} & \multirow{1}{*}{CLIP} & Cos Sim. & 42.40 & 57.22 & 67.98 \\
         & \multirow{1}{*}{CLIP-euc$^\dagger$} & Euc Ang$.*$ & 53.54 & 68.77 & 78.61   \\
         & \multirow{1}{*}{\cellcolor{blue!8} CLIP-hyp$^\dagger$} & \cellcolor{blue!8} Hyp Ang$.*$ & \cellcolor{blue!8} \textbf{53.87} & \cellcolor{blue!8} \textbf{69.13} & \cellcolor{blue!8} \textbf{78.85} \\
         \midrule
         \multirowcell{3}{OT Distance $\downarrow$} & \multirow{1}{*}{CLIP} & Cos Sim. & 6.25  & 7.89 & 8.76 \\
         & \multirow{1}{*}{CLIP-euc$^\dagger$} & Euc Ang$.*$ & \textbf{5.49} & \textbf{7.04}  & \textbf{8.05}  \\
         & \multirow{1}{*}{\cellcolor{blue!8} CLIP-hyp$^\dagger$} & \cellcolor{blue!8} Hyp Ang$.*$ & \cellcolor{blue!8} 5.56 & \cellcolor{blue!8} 7.13 & \cellcolor{blue!8} 8.13 \\
         \bottomrule
         
    \end{tabular}
    }
    \caption{
    \textbf{Out-of-domain part-based image retrieval evaluation on the LVIS dataset:
    }
    same-class (top) and part-based hierarchical retrieval (bottom).
    }
    \label{tab:unseen data}
    \vspace{-2ex}
\end{table}

\begin{table}[t]
    \centering 
    \resizebox{0.26\textwidth}{!}{
        \begin{tabular}{ccccc}
            \toprule
            Dataset  & CLIP  & CLIP-euc$^\dagger$  &  CLIP-hyp$^\dagger$  \\ 
            \midrule
            VOC  & 87.3 & 93.6 & \textbf{94.2}  \\
            COCO  & 63.6 & 73.4 & \textbf{73.9}
            \\
            \bottomrule
        \end{tabular}
    }
    \vspace{-1ex}
    \caption{
    \textbf{Zero-shot object detection using ground-truth boxes.} Image-only model with top-5 majority voting.
    }
    \label{tab:coco_voc}
    \vspace{-2ex}
\end{table}

\vspace{-1ex}
\section{Conclusion}
\label{sec:conlusion}

In this work, we introduce a new learning paradigm that effectively encodes user-defined visual hierarchies in hyperbolic space without requiring explicit hierarchical labels.
We present a concrete example of defining a part-based multi-level complex image hierarchy using object-level annotations and propose a contrastive loss in hyperbolic space to enforce pairwise entailment relationships.
Additionally, we introduce new evaluation metrics for hierarchical image retrieval.
Our experiments demonstrate our model effectively learns the predefined image hierarchy and goes beyond visual similarity.

{
    \small
    \bibliographystyle{ieeenat_fullname}
    \bibliography{main}
}

\clearpage
\appendix

\section*{APPENDICES}
\section{Hierarchy Tree}
\label{sec: Hierarchy Tree}
Our part-based image hierarchy framework is designed to be widely applicable to general image datasets with bounding box annotations.
In this section, we outline the key implementation involved in constructing hierarchy trees. 

We introduce a general method to generate dataset-specific ground truth hierarchy trees based on data statistics.
Following the approach outlined in Sec. 3.1 of the main paper, we begin by identifying bounding box pairs with substantial overlap.
In this work, we define a bounding box pair when at least 80\% of the smaller bounding box 
$b$ is contained within either the full image or a larger bounding box. We initially set the overlap threshold at 100\% and empirically adjusted it by evaluating the hierarchy’s validity using text labels. For the OpenImages dataset, 80\% was found to be a suitable threshold. This is a design choice and can be adapted for other datasets.

These pairs are then filtered based on two criteria: a \textit{frequency} threshold and a \textit{proportion} threshold.
For each pair, we record the \textit{frequency} of occurrences (\eg., bicycle-to-wheel relationships) and calculate the \textit{proportion} of instances where a child class appears within a parent class (\eg., the percentage of bicycle bounding boxes containing a wheel bounding box). 
Only pairs meeting both criteria, frequent occurrence and consistent labeling, are preserved.
We choose $\textit{frequency}=50$ and $\textit{proportion}=10\%$ in our experiments.

\begin{figure*}[t]
\centering\includegraphics[width=0.8\linewidth]{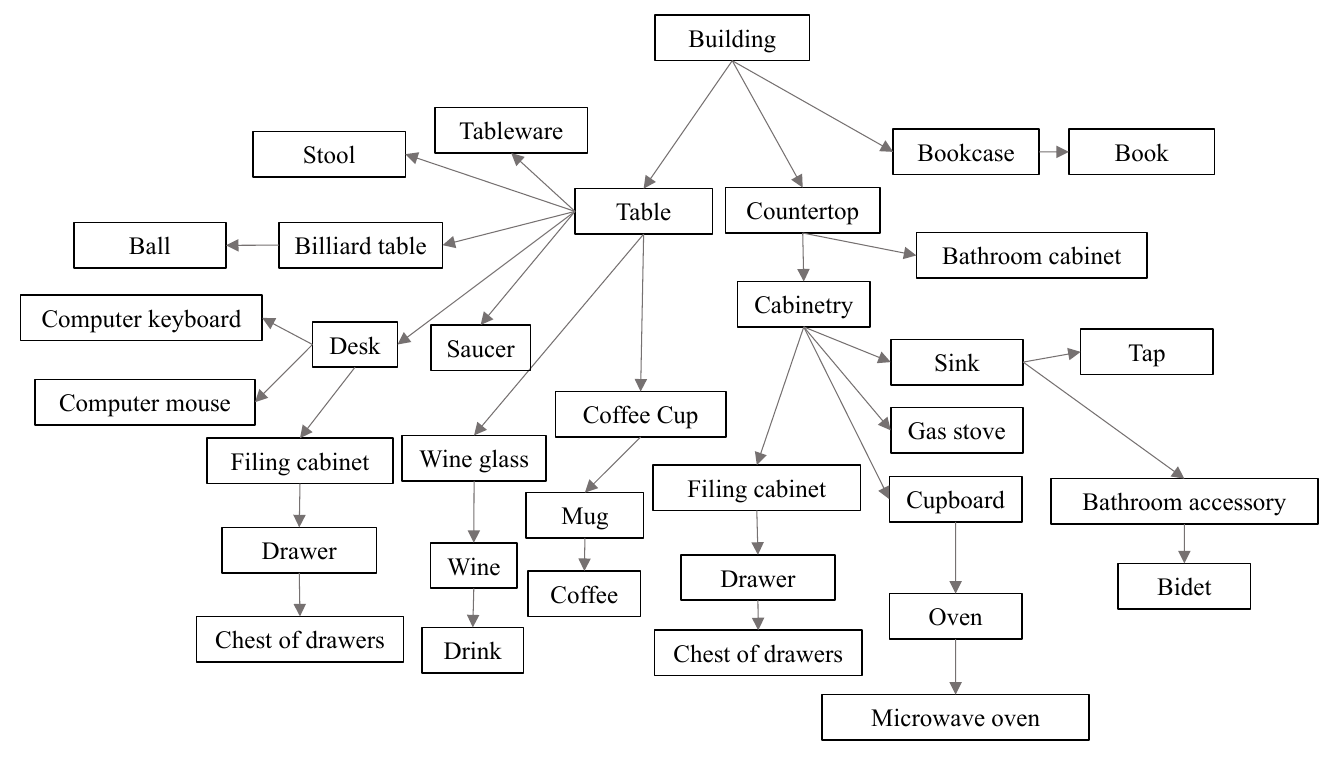}
\caption{\textbf{Example of a subset of hierarchical trees extracted from the OpenImages dataset.}
}
\label{fig:tree demo}
\end{figure*}

Once entailment pairs are established, they are organized into hierarchical trees (see examples in Fig.~\ref{fig:tree demo}).
In the evaluation of hierarchal image retrieval, the order of the hierarchy tree is essential:
for parent-to-child retrieval, 
lower-level concepts below the child in the hierarchy tree are considered correct, while for child-to-parent retrieval, higher-level concepts above the input classes are correct.

\section{Experiment Details}

\subsection{Hyperparameters and training details}
We employ the AdamW optimizer with parameters
$(\beta_1, \beta_2)=(0.9, 0.999)$ and a learning rate of $2 \times 10^{-5}$.
Training was conducted using 8 $\times$ A10G Nvidia GPUs. 
For each model, we used the largest batch size that fit in memory: CLIP ViT was trained with a total effective batch size of 640, and MoCo-v2 with a total effective batch size of 1984.
Each model was fine-tuned on \textit{HierOpenImages} dataset.
The embeddings are projected to dimension 128 in the final layer.
The hyperbolic model has a learnable curvature parameter.

During training, we filter out bounding boxes that occupy less than $1\%$ of the full image area, as well as pairs involving small bounding boxes labeled as `IsGroupOf' objects in the bounding box-to-bounding box relationships.
For data augmentation, we apply randomly horizontal flip ($20\%$), vertical flip($20\%$), rotate (degree {=15}), color jitter (brightness{=0.2}, contrast={0.2}, saturation{=0.2}, hue{=0.1}), Gaussian blur (kernel size{=5}, $\sigma=(0.3, 1.5)$), and then resize each image to $224\times224$.

\subsection{Part-based Image Retrieval}

\paragraph{Data.}
\textit{HierOpenImages} is built from the OpenImages dataset~\cite{kuznetsova2020open}, which originally contains approximately 1.9 million images, 14 million bounding boxes and 600 labels.
We create image-to-bounding box pairs, including one cross-image  bounding box sample for each bounding box class in the image, and bounding box to bounding box pairs where at least $80\%$ of the smaller bounding box is contained in the larger bounding box.

\begin{figure*}[t]
    \centering
\includegraphics[width=1\linewidth]{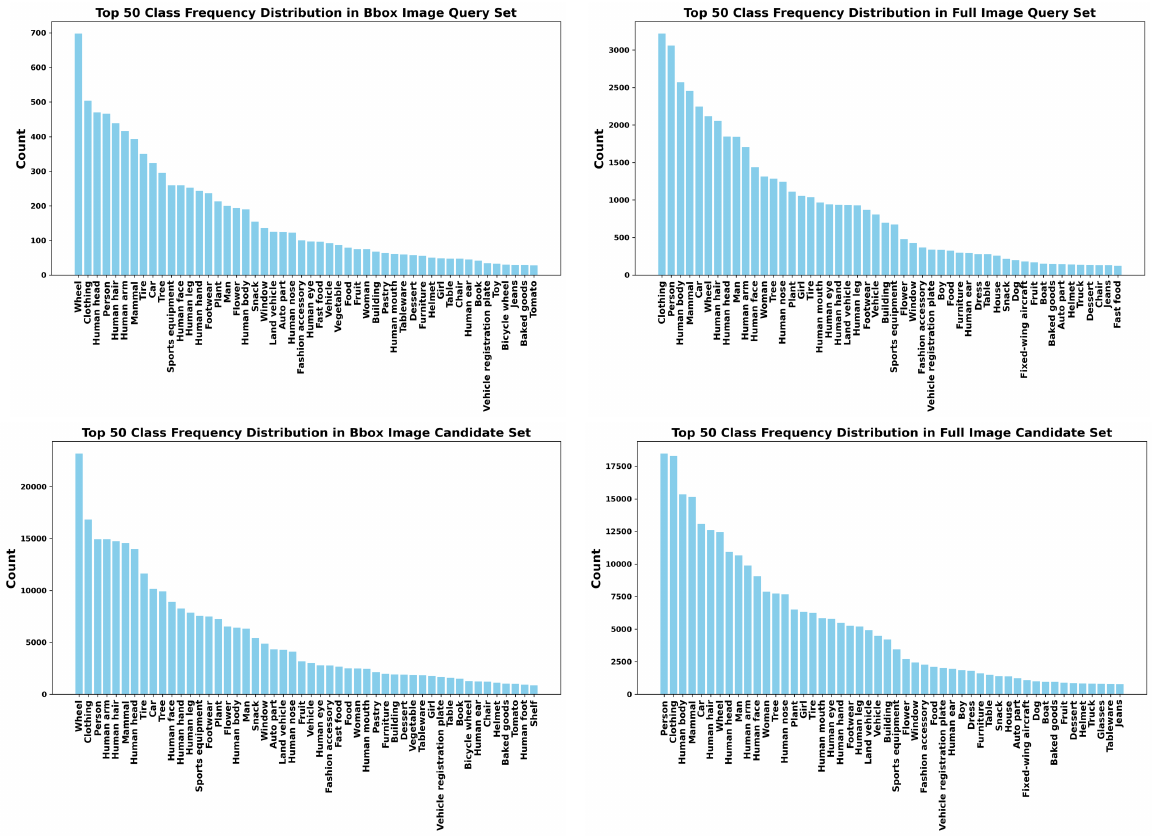}
    \caption{\textbf{Bounding box class distribution in the candidate and query sets.}}
    \label{fig:hist_class_bbox}
\end{figure*}

During part-based image retrieval evaluation,
we filter out bounding boxes for very small or large objects, and filter out noisy samples (\textit{e.g.,} wrong labels, highly occluded, \textit{etc}).
For part-based image retrieval, we randomly select a subset of 10,000 full images and 10,000 bounding box images as query images, using the entire filtered test set as candidates.
The top-50 class frequency distributions for both the query and candidate sets are shown in Fig.~\ref{fig:hist_class_bbox}.
Although the query and candidate set distributions are similar, the class distribution is highly imbalanced, highlighting the importance of hierarchical retrieval evaluation using combined precision-recall curves and OT distances (Sec.~3.3; see Fig.~4 and Table 2 in the main paper).

\paragraph{Gated Entailment Angle Metric.}
In this work, all entailment-angle thresholds are measured in radians. We use a threshold of 2 for MoCo and 2.5 for CLIP in small-k same-class classification, and a threshold of 1 for both models in large-k hierarchical retrieval.

\paragraph{Same-Class Retrieval.}
For part-to-full retrieval, a retrieval is considered correct if the retrieved full image contains the same object class as the query bounding box image.
For full-to-part retrieval, a retrieval is correct if the retrieved bounding box image corresponds to an object class within the query full image.

\paragraph{Hierarchical Retrieval.}
From a query parent image, correct child classes are all classes located at the lower level on the hierarchy tree of the labeled classes of the parent image (see examples in Fig.~\ref{fig:tree demo}).
For instance, 
when querying with a high-level full image, such as an image of a car, we expect to retrieve lower-level bounding boxes associated with the car, such as the car mirror, wheel or car plate \textit{etc}.
Conversely, when querying with a bounding box image, such as a wheel, we expect to retrieve various types of higher-level full images that include wheels, like cars, bicycles or cyclists \textit{etc}.

\paragraph{Hierarchal Retrieval Evaluation}
To compute the optimal transport (1-D Wasserstein) distance between the retrieved label distribution and the ground truth (Table~2 and Fig.~4 in the main paper),
we construct the ground truth distribution based on the frequency of each class (and its hierarchical classes) in the query parent image.
We count the occurrences of each class in the candidate set and build the ground truth distribution by normalizing the frequencies to sum to 1.
Similarly, the retrieval distribution is built by counting and normalizing retrieved class occurrences, 
and assigning any retrieved classes outside the ground truth hierarchy tree to an `others' class.
The two distributions are aligned by class order (ground truth distribution is zero in the `others' class), and the 1-D Wasserstein distance is computed using the \texttt{scipy} library.

Note that the Wasserstein distance has a closed-form formula for 1-D data.
If \( P \) and \( Q \) are represented as discrete empirical distributions (\eg., histograms or sorted samples of size \( n \)), let
$\{x_1, x_2, \ldots, x_n\}$ to be sorted values $P$, and $\{y_1, y_2, \ldots, y_n\}$ to be sorted values from $Q$,
then the 1-D Wasserstein distance is:
\[
W_p(P, Q) = \left( \frac{1}{n}\sum_{i=1}^n \left\| x_i - y_i \right\|^p \, \right)^{1/p},
\]
where $p$ refers to the order of the distance in the general p-Wasserstein metric.

In Table~2 of the main paper, we retrieve the Top-K results starting from a large K.
This is necessary because each image contains an average of 5.29 distinct classes and 61.5 classes across hierarchical trees, requiring a large number of retrievals to accurately evaluate the distribution.

\begin{figure*}[t]
\centering\includegraphics[width=0.9\linewidth]{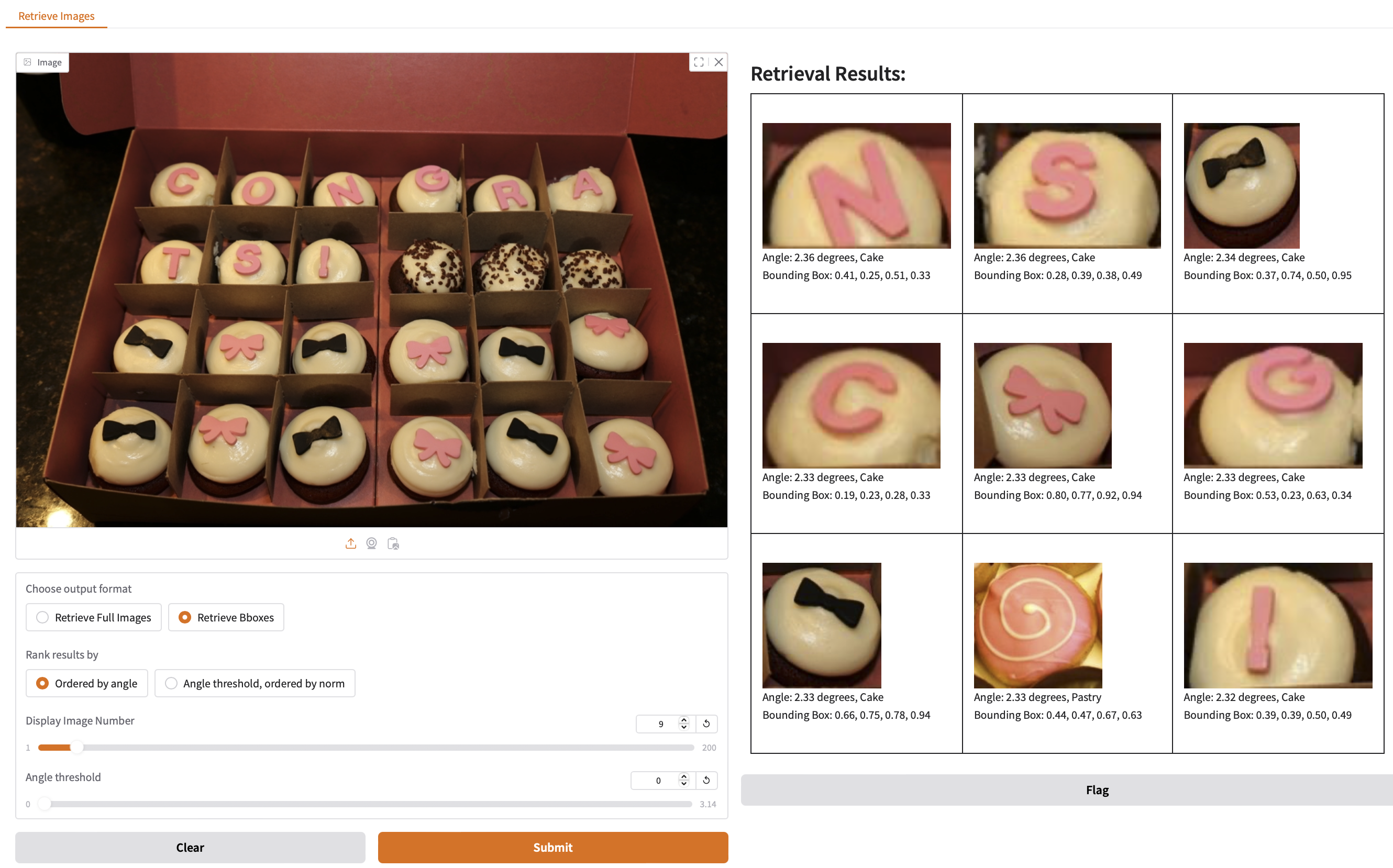}
\caption{\textbf{Example of our image retrieval interface built with Gradio~\cite{abid2019gradio}.} 
The interface supports image selection, upload and retrieves results ranked by user-defined modes. 
Its modular design allows for easy integration of additional functionalities.
}
\label{fig:gradio}
\end{figure*}

\paragraph{Gated Entailment Angle Metric.
}
\label{sec:metrics}
{\renewcommand{\arraystretch}{1.2} 
\begin{table}[t]
    \centering
    \small
    \vspace{-1ex}
    \resizebox{0.97\columnwidth}{!}{%
    \begin{tabular}{cc c cc cc}

        \toprule
         Vision  Encoder & Model & Metrics & \multicolumn{1}{c}{\cellcolor{orange!10} Top-5} & \multicolumn{1}{c}{\cellcolor{orange!20}Top-10} &  \multicolumn{1}{c}{\cellcolor{orange!30} Top-50} & \multicolumn{1}{c}{\cellcolor{orange!40}Top-100 }  \\ \midrule

        \rowcolor{Gray}\multicolumn{0}{c}{\textbf{Child-to-Parent }} & & & & & & \\
    
         \multirowcell{4}{ CLIP ViT} 

         & \multirow{1}{*}{CLIP-euc$^\dagger$} & Cos Sim.  & 75.20  & 74.32 & 71.81 & 70.21 \\
         & \multirow{1}{*}{
         CLIP-hyp$^\dagger$} &  Cos Sim.  &  76.82 & 75.53 & 72.46 & 70.50 \\   
         \cline{2-7}

         & \multirow{1}{*}{CLIP-euc$^\dagger$} & Euc Ang.  &  75.35 & 74.23 & 72.45 & 70.78  \\
         & \multirow{1}{*}{
         CLIP-hyp$^\dagger$} &  Hyp Ang. &  76.95 & 75.97 & 72.94 & 71.04 \\
         
         \cline{2-7}
         
         & \multirow{1}{*}{CLIP-euc$^\dagger$} & Euc Ang$.*$  &  75.63 & 74.65 & 72.25 &  70.52 \\
         & \multirow{1}{*}{
         CLIP-hyp$^\dagger$} &  Hyp Ang$.*$ & \textbf{77.28}  & \textbf{75.91} & \textbf{72.85} & \textbf{70.94}  \\

         \midrule
         \multirowcell{4}{ MoCo-v2} 
         & \multirow{1}{*}{MoCo-euc$^\dagger$} & Cos Sim.  &  67.93 & 67.67 & 66.20 & 65.04 \\
         & \multirow{1}{*}{
         MoCo-hyp$^\dagger$} &  Cos Sim.  & \textbf{69.00} & \textbf{68.37} & \textbf{66.26}  & \textbf{64.96}  \\
         \cline{2-7}
         & \multirow{1}{*}{MoCo-euc$^\dagger$} & Euc Ang. & 47.00 & 46.64 & 43.36 & 41.00 \\
         & \multirow{1}{*}{ MoCo-hyp$^\dagger$} &  Hyp Ang. &  49.91 & 48.93 & 43.73 & 40.95  \\
         \cline{2-7}
         & \multirow{1}{*}{MoCo-euc$^\dagger$} & Euc Ang$.^*$ & 66.78 & 66.58 & 65.42 & 64.40 \\
         & \multirow{1}{*}{ MoCo-hyp$^\dagger$} &  Hyp Ang$.^*$ & 68.62 &  67.42 & 66.21 & 64.94 \\

        \midrule

        \rowcolor{Gray}\multicolumn{0}{c}{\textbf{Parent-to-Child }} & & & & & & \\
    
         \multirowcell{4}{ CLIP ViT}

         & \multirow{1}{*}{CLIP-euc$^\dagger$} & Cos Sim.  & 73.73  &  72.52 & 69.97 & 68.70 \\
         & \multirow{1}{*}{
         CLIP-hyp$^\dagger$} &  Cos Sim.  & 74.38  & 73.20  & 70.67 & 69.28 \\
         \cline{2-7}
         & \multirow{1}{*}{CLIP-euc$^\dagger$} & Euc Ang. & 72.59 & 71.58 & 69.49 & 68.46 \\
         & \multirow{1}{*}{ CLIP-hyp$^\dagger$} &  Hyp Ang.  & 73.80 & 72.89 & 70.67 & 69.52 \\
         \cline{2-7}
         & \multirow{1}{*}{CLIP-euc$^\dagger$} & Euc Ang$.*$ & 73.74 & 72.57 & 70.04 & 68.78 \\
         & \multirow{1}{*}{ CLIP-hyp$^\dagger$} &  Hyp Ang$.*$  & \textbf{74.48} &  \textbf{73.27} &  \textbf{70.78} &  \textbf{69.46} \\
         
         \midrule
         \multirowcell{4}{ MoCo-v2} 
         
         & \multirow{1}{*}{MoCo-euc$^\dagger$} & Cos Sim.  &  66.68 & 65.90  & 64.20 & 63.25 \\
         & \multirow{1}{*}{
         MoCo-hyp$^\dagger$} &  Cos Sim.  & 66.80  & 65.88  & 64.07 & 63.11 \\
         \cline{2-7}
         & \multirow{1}{*}{MoCo-euc$^\dagger$} & Euc Ang.  &  67.21 & 66.23 & 63.47 & 61.85  \\
         & \multirow{1}{*}{
         MoCo-hyp$^\dagger$} &  Hyp Ang. & 67.53 & 66.14 & 62.50 & 60.82 \\
         \cline{2-7}
         & \multirow{1}{*}{MoCo-euc$^\dagger$} & Euc Ang$.*$  &  68.87 & 67.94 & 65.64 & 64.42 \\
         & \multirow{1}{*}{
         MoCo-hyp$^\dagger$} &  Hyp Ang$.*$ & \textbf{69.50} & \textbf{68.14} & \textbf{65.67} & \textbf{64.46}  \\
        \midrule
        \midrule
    \end{tabular}
    }
    \caption{\textbf{Part-based same-class image retrieval evaluation using different metrics. 
    }
    Cosine similarity (Cos Sim.) is the pretrained metric, 
    entailment angle (Euc or Hyp Ang.) is the finetuned metric, and $*$ represents a gated entailment angle metric which combines the strengths of both.
    }
    \label{tab:cos_angle}
\end{table}
}

In Table~\ref{tab:cos_angle}, we report retrieval performance using three metrics for completeness: cosine similarity (the model's pretrained metric), entailment angle (the finetuned metric), and a gated entailment angle metric (Ang.$*$), which combines the strengths of both.

The gated approach first uses the entailment angle as a cone-shaped filter to retain only candidates that satisfy the desired hierarchical direction (\textit{e.g.}, retrieving a parent or child image) and discards those that violate the entailment relationship. Cosine similarity is then applied within this gated set to provide fine-grained ranking. This two-stage procedure ensures that retrieval respects semantic hierarchy while leveraging the pretrained model’s strong embedding geometry.

A key finding is that fine-tuning with the entailment-angle loss not only improves performance under the entailment angle metric itself but also enhances the model’s original pretrained cosine-similarity performance. This suggests that the fine-tuning introduces meaningful hierarchical structure into the representation space in a way that strengthens the model’s pretrained retrieval capability. This observation aligns with the improvements observed in the zero-shot retrieval or classification results on LVIS, VOC, and COCO in the main paper.

Our intuition is that for models with strong pretrained priors or similar training setups (\textit{e.g.}, MoCo-v2, which is trained on image-only data with invariance to random crops), entailment angle fine-tuning primarily reshapes the coarse organization of the embedding space, aligning representations with the angle-based entailment structure. However, because these models inherit strong inductive priors from massive pretraining, small finetuning cannot fully reorient the fine-grained similarity relationships.
The gated metric effectively leverages these complementary strengths. The entailment angle enforces the correct hierarchical region, while cosine similarity provides accurate ranking within that region. By combining directional gating with pretrained metric ordering, the hybrid approach balances the new hierarchical signal with the stability of the pretrained embedding, yielding a more robust and reliable retrieval strategy.

\paragraph{Image Retrieval Interface via Gradio.
}

We built our image retrieval interface using Gradio~\cite{abid2019gradio}, as shown in Fig.~\ref{fig:gradio}. Input images can be selected from a linked image folder, where thumbnail images are displayed in an image gallery, or they can be directly uploaded by users. The retrieval results can be sorted by the hyperbolic angle relative to the input image or filtered using a user-defined threshold value, after which the results are ordered by their embedding norms. Additional functionalities can be easily integrated into the current pipeline.

\begin{figure*}[h!]
    \centering 
    \includegraphics[width=0.9\linewidth]{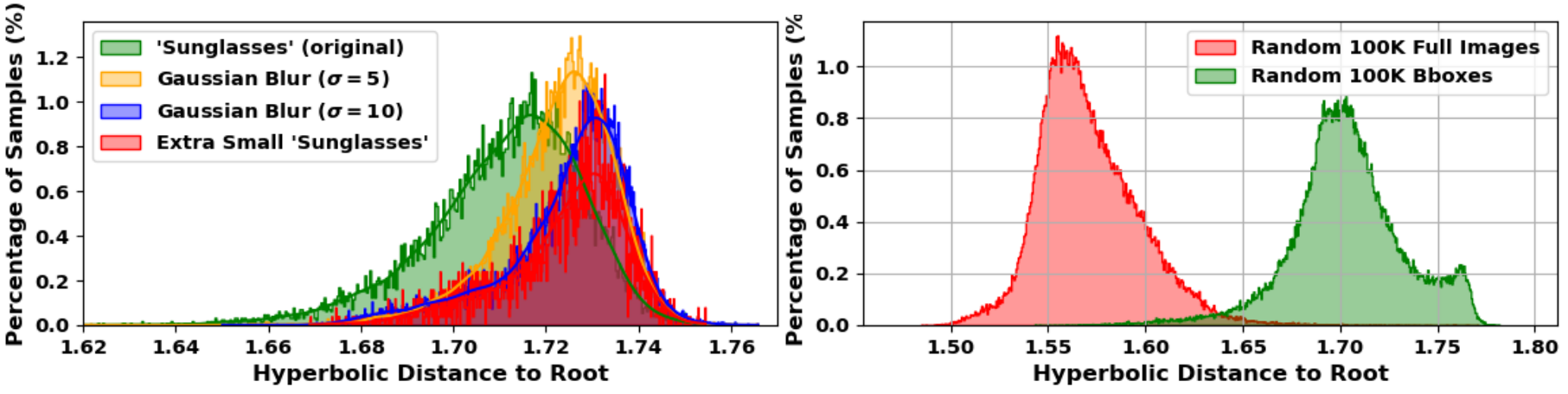} 
    \caption{\textbf{Example of the model's emergent structure.}
    In the left plot, the percentage of samples of the original \textit{Sunglasses} crops are shown in dark green.
    Applying Gaussian blur or selecting extremely small crops shifts the embeddings further from the root. 
    The right figure illustrates a clear separation between full-image and part-image embeddings.
    }
    \label{fig:dist}
\end{figure*}

\subsection{Generalization Evaluation}

\paragraph{LVIS Dataset.} We evaluate the generalization capability of our model on the out-of-domain LVIS dataset~\cite{gupta2019lvis}, which is designed for large-scale ($\sim$1.2M bounding boxes) long-tail instance classification and segmentation.
It has a highly imbalanced distribution of 1,203 object categories and contains 897 object categories that are absent from OpenImages~\cite{kuznetsova2020open}.
Here are some examples of unseen hierarchies in LVIS~\cite{gupta2019lvis} but not in OpenImges~\cite{kuznetsova2020open}: \{table $\rightarrow$ tablecloth $\rightarrow$ ashtray $\rightarrow$ cigarette\}, 
\{backpack $\rightarrow$ strap $\rightarrow$ belt buckle\},
\{toy $\rightarrow$ teddy bear $\rightarrow$ thread $\rightarrow$ bobbin\}, \{sofa $\rightarrow$ blanket $\rightarrow$ quilt $\rightarrow$ bedspread\}.
The full list of these categories can be found in the appendix.
Only display the first class of synonyms.

We construct the hierarchical evaluation set from the validation set of the LVIS dataset~\cite{gupta2019lvis}, following the similar process as constructing~\textit{HierOpenImages}.
To construct the ground truth hierarchy tree, we use the pipeline as described in Sec.~\ref{sec: Hierarchy Tree}.
We empirically choose parameter $\textit{frequency}=5$ and $\textit{proportion}=5\%$ in our experiments. 
This adjustment is necessary because the LVIS dataset~\cite{gupta2019lvis} has very unbalanced classes.

\paragraph{Emergent structure.}
 We observe that high-frequency, ambiguous image crops are naturally pushed toward the boundary of hyperbolic space, where more angular entailment constraints can be better satisfied. 
This aligns with the exponentially growing volume near the boundary, and it is an efficient solution for representing diverse, overlapping semantics.
In Figure~\ref{fig:dist}, in the left plot, the percentage of samples of the original \textit{Sunglasses} crops are shown in dark green.
Adding Gaussian blur or selecting extremely small crops shifts their embeddings further from the root.
The small crops (red curve) are not included in other curves.
The right plot shows clear separation between full and part image embeddings.

\begin{table}[t]
    \centering
    \small
    \resizebox{\columnwidth}{!}{%
    \begin{tabular}{cc cc cc}

        \toprule
         Model & Cross-Image  &  \multicolumn{1}{c}{\cellcolor{orange!10} Top-5} & \multicolumn{1}{c}{\cellcolor{orange!20}Top-10} &  \multicolumn{1}{c}{\cellcolor{orange!30} Top-50} & \multicolumn{1}{c}{\cellcolor{orange!40}Top-100}  \\ \midrule

        \rowcolor{Gray}\multicolumn{0}{c}{\textbf{Child-to-Parent}} & & & & & \\

         CLIP-hyp$^\dagger$ &  \cmark & \textbf{77.28}  & \textbf{75.91} & \textbf{72.85} &  \textbf{70.94}   \\
         CLIP-euc$^\dagger$ & \cmark &  75.63 & 74.65 & 72.25 &  70.52  \\
         CLIP-hyp$^\dagger$ &  \xmark & 73.56 & 71.57 & 68.25 & 66.60 \\
         CLIP-euc$^\dagger$ & \xmark  & 73.32 & 71.26 & 67.33 & 65.67 \\

         CLIP & - & 53.04 & 51.69 & 47.87 & 45.79 \\
          \rowcolor{Gray}{\textbf{Parent-to-Child}}  & & & & &   \\

          CLIP-hyp$^\dagger$ & \cmark  &  \textbf{74.48} & \textbf{73.27} & \textbf{70.78} & \textbf{69.46}  \\
          CLIP-euc$^\dagger$ & \cmark &  73.74 & 72.57 & 70.04 & 68.78  \\
          CLIP-hyp$^\dagger$ & \xmark &  71.87 & 69.58 & 65.31 & 63.95 \\
          CLIP-euc$^\dagger$ & \xmark &  71.61 & 69.00 & 64.69 & 63.28 \\
         CLIP & - & 69.94  & 67.74 & 62.50  & 60.21 \\

         \bottomrule
         
    \end{tabular}
    }
    \caption{\textbf{Part-based same-class image retrieval evaluation.} 
    Cross-image \cmark indicates models fine-tuned on entailment relationships both within and across images at the category level, while \xmark~represents models fine-tuned without cross-image sampling.
    The evaluation setup is the same as Table 1 in the main paper.
    The best results are marked in bold.
    }
    \label{tab:same-class_1}
\end{table}
\begin{table}[t]
    \centering
    \small
    \vspace{-1ex}
    \resizebox{1\columnwidth}{!}{%
    \begin{tabular}{cc c cc c}

        \toprule
         Metrics & Model & Cross-Image &  \multicolumn{1}{c}{ Top-150k} & \multicolumn{1}{c}{Top-200k} &  \multicolumn{1}{c}{ Top-250k}  \\ \midrule
         
         & \multirow{1}{*}{CLIP-hyp$^\dagger$}  & \checkmark & \textbf{77.00} & \textbf{85.75} & \textbf{91.89}  \\
         & \multirow{1}{*}{CLIP-euc$^\dagger$} & \checkmark & 76.46 & 85.08 & 91.38  \\
         & \multirow{1}{*}{CLIP-hyp$^\dagger$}  & \xmark & 75.66 & 84.51 & 91.13 \\
         & \multirow{1}{*}{CLIP-euc$^\dagger$} & \xmark & 75.22 & 84.15 & 91.10 \\
         \multirowcell{-5}{Recall $\%\uparrow$} & \multirow{1}{*}{CLIP} & - & 66.63 & 77.15  & 86.57 \\
         \midrule
         & \multirow{1}{*}{CLIP-hyp$^\dagger$} & \checkmark & \textbf{14.96} & \textbf{17.57} & \textbf{20.76}   \\
         & \multirow{1}{*}{CLIP-euc$^\dagger$} & \checkmark & 15.65 & 18.15 & 21.09   \\
         & \multirow{1}{*}{CLIP-hyp$^\dagger$}  & \xmark & 15.98 & 18.27 & 21.03 \\
         & \multirow{1}{*}{CLIP-euc$^\dagger$} & \xmark & 16.19 & 18.46 & 21.02 \\
         \multirowcell{-5}{OT Distance $\downarrow$} & \multirow{1}{*}{CLIP}  & - & 21.31 & 22.74 & 23.79 \\

         \bottomrule
         
    \end{tabular}
    }
    \caption{\textbf{Part-based hierarchical evaluation of parent-to-child image retrieval.}
    The OT distance is defined in Sec.~3.3 (main paper), and the evaluation setup follows Table 2 (main paper).
    The best results are marked in bold.
    }
    \label{tab:hier_2}
    \vspace{-2ex}
\end{table}

\section{Ablation Studies}

In this ablation study, we evaluate the impact of adding cross-image scene-to-object samples.
We fine-tuned the CLIP ViT model solely on hierarchical part-based entailment data within individual images (entailment pairs with high visual similarity), excluding any cross-image image-to-bounding-box samples.
The detailed results for same-class retrieval and hierarchical retrieval are presented in Table~\ref{tab:same-class_1}–\ref{tab:hier_2}, with the best scores highlighted in bold. Overall, the results clearly show that incorporating cross-image samples further improves image retrieval performance.

\section{More Qualitative Results}

\begin{figure*}[t]
\centering\includegraphics[width=0.85\linewidth]{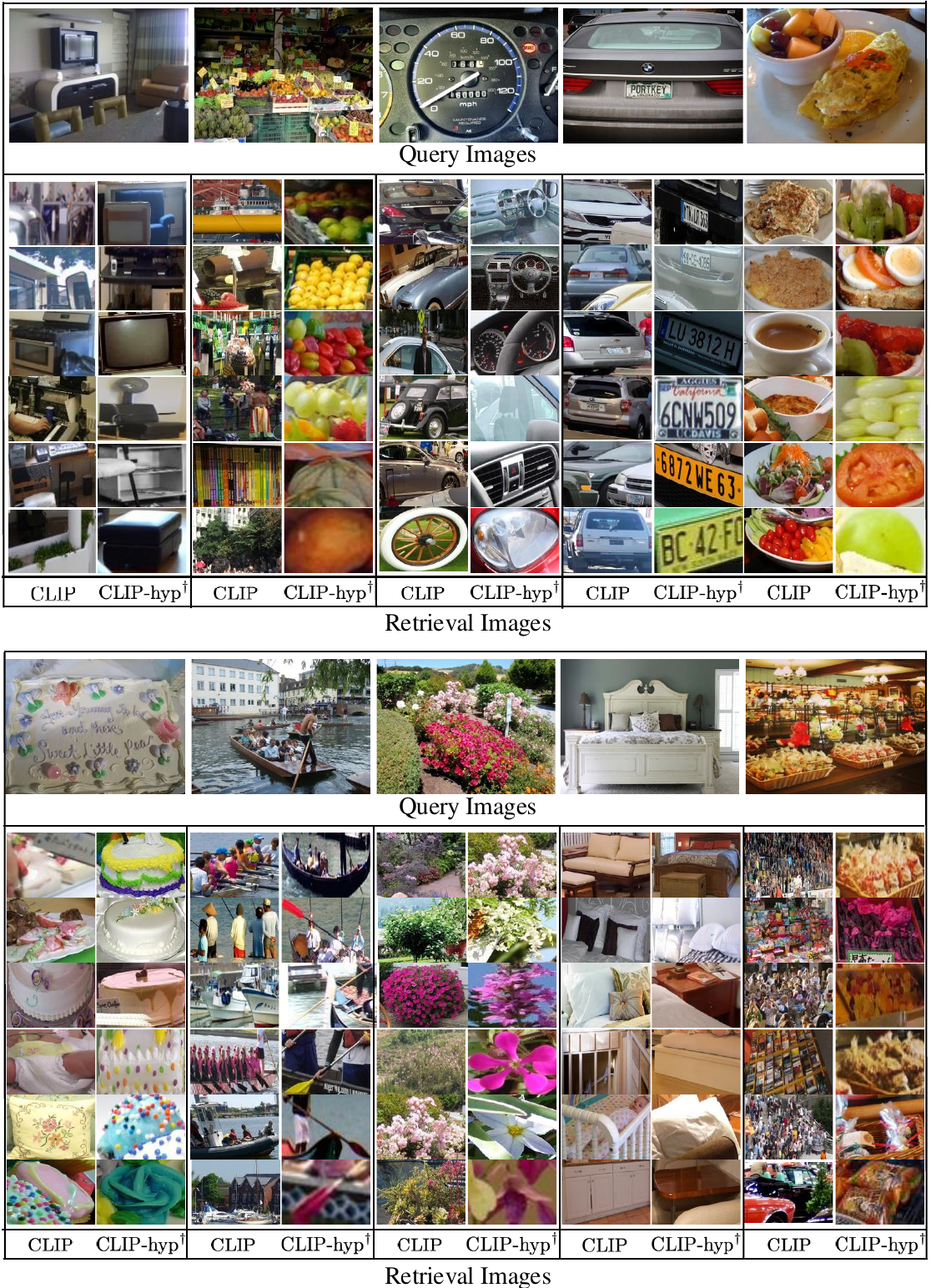}
\caption{\textbf{Example of parent-to-child retrieval using CLIP ViT and our CLIP-hyp$^\dagger$ model.} 
	Results are ordered by ascending embedding norms.
	Our model retrieves images matching the predefined {\em scene-object-part} hierarchy, placing high-level objects near the origin (\eg, group of fruits $\rightarrow$ single fruits), and grouping semantically related but visually distinct objects (\eg, chairs \& TVs).
    All retrieved bounding box images are scaled to the same ratio.
}
\label{fig:more demo}
\end{figure*}

More qualitative parent-to-child retrieval results are shown in Fig.~\ref{fig:more demo} to visualize the latent space.
Bounding box images are filtered by angle (CLIP-hyp$^\dagger$) or cosine similarity (CLIP ViT model) thresholds and sorted by increasing embedding norms.
Our hyperbolic model retrieves diverse and visually distinct lower-level objects related to the query images, organized according to the predefined \textit{scene-object-part} hierarchy in the embedding space.

\clearpage
\begin{table}[htbp]
    \small
    \centering
    \renewcommand{\arraystretch}{1.2} 
    \begin{tabular}{|p{0.15\textwidth}|p{0.15\textwidth}|p{0.15\textwidth}|p{0.15\textwidth}|p{0.15\textwidth}|p{0.15\textwidth}|}
    \multicolumn{6}{l}{\Huge \textbf{Appendix}} \\
    \multicolumn{6}{l}{} \\
    \multicolumn{6}{l}{\large \textbf{Object classes in LVIS dataset but not in OpenImage dataset}} \\
    \multicolumn{6}{l}{} \\
    \hline
    aerosol can & air conditioner & alcohol & alligator & almond & amplifier \\ \hline
anklet & antenna & applesauce & apricot & apron & aquarium \\ \hline
arctic & armband & armchair & armoire & armor & trash can \\ \hline
ashtray & asparagus & atomizer & avocado & award & awning \\ \hline
baboon & baby buggy & basketball backboard & bagpipe & baguet & bait \\ \hline
ballet skirt & bamboo & Band Aid & bandage & bandanna & banner \\ \hline
barbell & barrette & barrow & baseball base & baseball & baseball cap \\ \hline
basket & basketball & bass horn & bat & bath mat & bath towel \\ \hline
bathrobe & batter & battery & beachball & bead & bean curd \\ \hline
beanbag & beanie & bedpan & bedspread & cow & beef \\ \hline
beeper & beer bottle & beer can & bell & belt buckle & beret \\ \hline
bib & Bible & visor & binder & birdfeeder & birdbath \\ \hline
birdcage & birdhouse & birthday cake & birthday card & pirate flag & black sheep \\ \hline
blackberry & blackboard & blanket & blazer & blimp & blinker \\ \hline
blouse & blueberry & gameboard & bob & bobbin & bobby pin \\ \hline
boiled egg & bolo tie & deadbolt & bolt & bonnet & booklet \\ \hline
bookmark & boom microphone & bouquet & bow & bow & bow-tie \\ \hline
pipe bowl & bowler hat & bowling ball & boxing glove & suspenders & bracelet \\ \hline
brass plaque & bread-bin & breechcloth & bridal gown & broach & broom \\ \hline
brownie & brussels sprouts & bubble gum & bucket & horse buggy & horned cow \\ \hline
bulldog & bulldozer & bullet train & bulletin board & bulletproof vest & bullhorn \\ \hline
bun & bunk bed & buoy & business card & butter & button \\ \hline
cabana & cabin car & cabinet & locker & calendar & calf \\ \hline
camcorder & camera lens & camper & candle holder & candy bar & candy cane \\ \hline
walking cane & canister & canteen & cap & bottle cap & cape \\ \hline
cappuccino & railcar & elevator car & car battery & identity card & card \\ \hline
cardigan & cargo ship & carnation & horse carriage & tote bag & carton \\ \hline
cash register & casserole & cassette & cast & cauliflower & cayenne \\ \hline
CD player & celery & chain mail & chaise longue & chalice & chandelier \\ \hline
chap & checkbook & checkerboard & cherry & chessboard & chickpea \\ \hline
chili & chinaware & crisp & poker chip & chocolate bar & chocolate cake \\ \hline
chocolate milk & chocolate mousse & choker & chopstick & slide & cider \\ \hline
cigar box & cigarette & cigarette case & cistern & clarinet & clasp \\ \hline
cleansing agent & cleat & clementine & clip & clipboard & clippers \\ \hline
cloak & clock tower & clothes hamper & clothespin & clutch bag & coaster \\ \hline
coat hanger & coatrack & cock & cockroach & cocoa & coffee maker \\ \hline

\end{tabular}
\end{table}

\clearpage
\begin{table}[htbp]
    \small
    \centering
    \renewcommand{\arraystretch}{1.2} 
    \begin{tabular}{|p{0.15\textwidth}|p{0.15\textwidth}|p{0.15\textwidth}|p{0.15\textwidth}|p{0.15\textwidth}|p{0.15\textwidth}|}
    \hline
    coffeepot & coil & colander & coleslaw & coloring material & combination lock \\ \hline
    pacifier & comic book & compass & condiment & cone & control \\ \hline
convertible & cooker & cooking utensil & cooler & cork & corkboard \\ \hline
corkscrew & edible corn & cornbread & cornice & cornmeal & corset \\ \hline
costume & cougar & coverall & cowbell & crabmeat & cracker \\ \hline
crape & crate & crayon & cream pitcher & crib & crock pot \\ \hline
crossbar & crouton & crow & crowbar & crucifix & cruise ship \\ \hline
police cruiser & crumb & cub & cube & cufflink & cup \\ \hline
trophy cup & cupcake & hair curler & curling iron & cushion & cylinder \\ \hline
cymbal & dalmatian & dartboard & date & deck chair & dental floss \\ \hline
detergent & diary & dinghy & dining table & tux & dish \\ \hline
dish antenna & dishrag & dishtowel & dishwasher detergent & dispenser & diving board \\ \hline
Dixie cup & dog collar & dollar & dollhouse & domestic ass & doorknob \\ \hline
doormat & dove & underdrawers & dress hat & dress suit & dresser \\ \hline
drill & drone & dropper & drumstick & duckling & duct tape \\ \hline
duffel bag & dumpster & dustpan & earphone & earplug & earring \\ \hline
easel & eclair & eel & egg & egg roll & egg yolk \\ \hline
eggbeater & eggplant & electric chair & elk & escargot & eyepatch \\ \hline
fan & ferret & Ferris wheel & ferry & fig & fighter jet \\ \hline
figurine & file & fire alarm & fire engine & fire extinguisher & fire hose \\ \hline
first-aid kit & fishbowl & fishing rod & flagpole & flamingo & flannel \\ \hline
flap & flash & fleece & flip-flop & flipper & flower arrangement \\ \hline
flute glass & foal & folding chair & footstool & forklift & freight car \\ \hline
French toast & freshener & frisbee & fruit juice & fudge & funnel \\ \hline
futon & gag & garbage & garbage truck & garden hose & gargle \\ \hline
gargoyle & garlic & gasmask & gazelle & gelatin & gemstone \\ \hline
generator & gift wrap & ginger & cincture & glass & globe \\ \hline
golf club & golfcart & gorilla & gourd & grater & gravestone \\ \hline
gravy boat & green bean & green onion & griddle & grill & grits \\ \hline
grizzly & grocery bag & gull & gun & hairbrush & hairnet \\ \hline
hairpin & halter top & ham & hammock & hamper & hand glass \\ \hline
hand towel & handcart & handcuff & handkerchief & handle & handsaw \\ \hline
hardback book & harmonium & hatbox & veil & headband & headboard \\ \hline
headlight & headscarf & headset & headstall & heart & heron \\ \hline
highchair & hinge & hockey stick & home plate & honey & fume hood \\ \hline
hook & hookah & hornet & hose & hot-air balloon & hotplate \\ \hline
hot sauce & hourglass & houseboat & hummingbird & hummus & icecream \\ \hline
popsicle & ice maker & ice pack & ice skate & igniter & inhaler \\ \hline
iron & ironing board & jam & jar & jean & jeep \\ \hline
jelly bean & jersey & jet plane & jewel & jewelry & joystick \\ \hline

\end{tabular}
\end{table}

\clearpage
\begin{table}[htbp]
    \small
    \centering
    \renewcommand{\arraystretch}{1.2} 
    \begin{tabular}{|p{0.15\textwidth}|p{0.15\textwidth}|p{0.15\textwidth}|p{0.15\textwidth}|p{0.15\textwidth}|p{0.15\textwidth}|}
    \hline
    jumpsuit & kayak & keg & kennel & key & keycard \\ \hline
    kilt & kimono & kitchen sink & kitchen table & kitten & kiwi fruit \\ \hline
knee pad & knitting needle & knob & knocker & lab coat & lamb \\ \hline
lamb-chop & lamppost & lampshade & lanyard & laptop computer & lasagna \\ \hline
latch & lawn mower & leather & legging & Lego & legume \\ \hline
lemonade & lettuce & license plate & life buoy & life jacket & lightbulb \\ \hline
lightning rod & lime & lip balm & liquor & log & lollipop \\ \hline
speaker & machine gun & magazine & magnet & mail slot & mailbox \\ \hline
mallard & mallet & mammoth & manatee & mandarin orange & manager \\ \hline
manhole & map & marker & martini & mascot & mashed potato \\ \hline
masher & mask & mast & mat & matchbox & mattress \\ \hline
meatball & medicine & melon & microscope & milestone & milk can \\ \hline
milkshake & minivan & mint candy & mitten & money & monitor \\ \hline
motor & motor scooter & motor vehicle & mound & mousepad & music stool \\ \hline
nailfile & napkin & neckerchief & needle & nest & newspaper \\ \hline
newsstand & nightshirt & nosebag & noseband & notebook & notepad \\ \hline
nut & nutcracker & oar & octopus & octopus & oil lamp \\ \hline
olive oil & omelet & onion & orange juice & ottoman & overalls \\ \hline
packet & inkpad & pad & padlock & paintbrush & painting \\ \hline
pajamas & palette & pan & pan & pantyhose & papaya \\ \hline
paper plate & paperback book & paperweight & parakeet & parasail & parasol \\ \hline
parchment & parka & passenger car & passenger ship & passport & patty \\ \hline
pea & peanut butter & peeler & wooden leg & pegboard & pelican \\ \hline
pencil & pendulum & pennant & penny & pepper & pepper mill \\ \hline
persimmon & pet & pew & phonebook & phonograph record & pickle \\ \hline
pickup truck & pie & pigeon & piggy bank & pin & pinecone \\ \hline
ping-pong ball & pinwheel & tobacco pipe & pipe & pita & pitcher \\ \hline
pitchfork & place mat & playpen & pliers & plow & plume \\ \hline
pocket watch & pocketknife & poker & pole & polo shirt & poncho \\ \hline
pony & pop & postbox & postcard & pot & potholder \\ \hline
pottery & pouch & power shovel & projector & propeller & prune \\ \hline
pudding & puffer & puffin & pug-dog & puncher & puppet \\ \hline
puppy & quesadilla & quiche & quilt & race car & radar \\ \hline
radiator & radio receiver & raft & rag doll & raincoat & ram \\ \hline
raspberry & rat & razorblade & reamer & rearview mirror & receipt \\ \hline
recliner & record player & reflector & rib & ring & river boat \\ \hline
road map & robe & rocking chair & rodent & roller skate & Rollerblade \\ \hline
rolling pin & root beer & router & rubber band & runner & saddle \\ \hline
saddle blanket & saddlebag & safety pin & sail & salad plate & salami \\ \hline
salmon & salmon & salsa & saltshaker & satchel & saucepan \\ \hline
sausage & sawhorse & scarecrow & school bus & scraper & scrubbing brush \\ \hline

\end{tabular}
\end{table}

\clearpage
\begin{table}[htbp]
    \small
    \centering
    \renewcommand{\arraystretch}{1.2} 
    \begin{tabular}{|p{0.15\textwidth}|p{0.15\textwidth}|p{0.15\textwidth}|p{0.15\textwidth}|p{0.15\textwidth}|p{0.15\textwidth}|}
    \hline
    seabird & seaplane & seashell & shaker & shampoo & sharpener \\ \hline
    Sharpie & shaver & shaving cream & shawl & shears & shepherd dog \\ \hline
sherbert & shield & shoe & shopping bag & shopping cart & shot glass \\ \hline
shoulder bag & shovel & shower head & shower cap & shower curtain & shredder \\ \hline
signboard & silo & skewer & ski boot & ski parka & ski pole \\ \hline
skullcap & sled & sleeping bag & sling & slipper & smoothie \\ \hline
soap & soccer ball & softball & solar array & soup & soup bowl \\ \hline
soupspoon & sour cream & soya milk & space shuttle & sparkler & spear \\ \hline
crawfish & sponge & sportswear & spotlight & stagecoach & statue \\ \hline
steak & steak knife & steering wheel & stepladder & step stool & stereo \\ \hline
stew & stirrer & stirrup & brake light & stove & strainer \\ \hline
strap & street sign & streetlight & string cheese & stylus & subwoofer \\ \hline
sugar bowl & sugarcane & sunflower & sunhat & mop & sweat pants \\ \hline
sweatband & sweater & sweatshirt & sweet potato & Tabasco sauce & table-tennis table \\ \hline
table lamp & tablecloth & tachometer & tag & taillight & tambourine \\ \hline
army tank & tank top & tape & tape measure & tapestry & tarp \\ \hline
tartan & tassel & tea bag & teacup & teakettle & telephone booth \\ \hline
telephone pole & telephoto lens & television camera & television set & tequila & thermometer \\ \hline
thermos bottle & thermostat & thimble & thread & thumbtack & tights \\ \hline
timer & tinfoil & tinsel & tissue paper & toast & toaster oven \\ \hline
tongs & toolbox & toothpaste & toothpick & cover & tortilla \\ \hline
tow truck & towel rack & tractor & dirt bike & trailer truck & trampoline \\ \hline
tray & trench coat & triangle & tricycle & truffle & trunk \\ \hline
vat & turban & turnip & turtleneck & typewriter & underwear \\ \hline
urinal & urn & vacuum cleaner & vending machine & vent & vest \\ \hline
videotape & vinegar & vodka & volleyball & vulture & wagon \\ \hline
wagon wheel & walking stick & wall socket & wallet & walrus & washbasin \\ \hline
water bottle & water cooler & water heater & water jug & water gun & water ski \\ \hline
water tower & watering can & weathervane & webcam & wedding cake & wedding ring \\ \hline
wet suit & whipped cream & whistle & wig & wind chime & windmill \\ \hline
window box & windshield wiper & windsock & wine bottle & wine bucket & wineglass \\ \hline
blinder & wolf & wooden spoon & wreath & wristband & wristlet \\ \hline
yacht & yogurt & yoke & & & \\ \hline

\end{tabular}
\end{table}
\end{document}